\documentclass[lettersize,journal]{IEEEtran}
\usepackage{amsmath,amsfonts}
\usepackage{algorithmicx}
\usepackage{algorithm}
\usepackage{array}
\usepackage[caption=false,font=normalsize,labelfont=sf,textfont=sf]{subfig}
\usepackage{textcomp}
\usepackage{stfloats}
\usepackage{url}
\usepackage{verbatim}
\usepackage{graphicx}
\usepackage{cite}

\usepackage{bm}
\usepackage{color}
\usepackage{booktabs}
\newcommand{\red}[1]{{\color{red} #1}}

\usepackage{multirow}
\usepackage{enumitem}

\usepackage{listings}
\usepackage{diagbox}
\usepackage{etoolbox}

\usepackage{color, colortbl}
\usepackage[dvipsnames]{xcolor}
\definecolor{DnCBG}{rgb}{0.9, 0.9, 1.}

\newcommand{\gray}[1]{\textcolor{gray}{#1}}
\newcommand{\green}[1]{\textcolor[RGB]{96,177,87}{#1}}

\newcommand{\EE}{\mathbb{E}}
\newlength\savewidth\newcommand\shline{\noalign{\global\savewidth\arrayrulewidth\global\arrayrulewidth 1pt}\hline\noalign{\global\arrayrulewidth\savewidth}}

\usepackage[pagebackref=true,breaklinks=true,letterpaper=true,colorlinks,bookmarks=false]{hyperref}

\begin{document}

\title{Efficient Masked Autoencoders with Self-Consistency}

\author{
    Zhaowen Li,
    Yousong Zhu$^{\star}$,
    Zhiyang Chen,
    Wei Li,
    Rui Zhao,
    Chaoyang Zhao,
    Ming Tang,~\IEEEmembership{Member,~IEEE}
    Jinqiao Wang$^{\star}$,~\IEEEmembership{Member,~IEEE},
\IEEEcompsocitemizethanks{\IEEEcompsocthanksitem Zhaowen Li, Yousong Zhu, Zhiyang Chen, Chaoyang Zhao, and Jinqiao Wang are with the Foundation Model Research Center, Institute of Automation, Chinese Academy of Sciences, Beijing 100864, China. Zhaowen Li, Zhiyang Chen, and Jinqiao Wang are also with the School of Artificial Intelligence, University of Chinese Academy of Sciences, Beijing 100190, China. Jinqiao Wang is also Wuhan AI Research, Wuhan, China, and also with the Peng Cheng Laboratory, Shenzhen 518066, China.  Wei Li and Rui Zhao are with SenseTime Smart City Group, Beijing 100190, China(e-mail: \{zhaowen.li, yousong.zhu, zhiyang.chen, chaoyang.zhao, tangm, jqwang\}@nlpr.ia.ac.cn, \{liwei1, zhaorui\}@sensetime.com).
}
\thanks{Corresponding author: Yousong Zhu, Jinqiao Wang.}
\thanks{Manuscript received July 28, 2023.}}

\markboth{Submitted to IEEE TRANSACTIONS ON Pattern Analysis and Machine Intelligence}%
{Li \MakeLowercase{\textit{et al.}}: Efficient Masked Autoencoders with Self-Consistency}


\maketitle

\begin{abstract}
 Inspired by the masked language modeling (MLM) in natural language processing tasks, the masked image modeling (MIM) has been recognized as a strong self-supervised pre-training method in computer vision. However, the high random mask ratio of MIM results in two serious problems: 1) the inadequate data utilization of images within each iteration brings prolonged pre-training, and 2) the high inconsistency of predictions results in unreliable generations, $i.e.$, the prediction of the identical patch may be inconsistent in different mask rounds, leading to divergent semantics in the ultimately generated outcomes. To tackle these problems, we propose the efficient masked autoencoders with self-consistency (EMAE) to improve the pre-training efficiency and increase the consistency of MIM. In particular, we present a parallel mask strategy that divides the image into K non-overlapping parts, each of which is generated by a random mask with the same mask ratio. Then the MIM task is conducted parallelly on all parts in an iteration and the model minimizes the loss between the predictions and the masked patches. Besides, we design the self-consistency learning to further maintain the consistency of predictions of overlapping masked patches among parts. Overall, our method is able to exploit the data more efficiently and obtains reliable representations. Experiments on ImageNet show that EMAE achieves the best performance on ViT-Large with only $13\%$ of MAE pre-training time using NVIDIA A100 GPUs. After pre-training on diverse datasets, EMAE consistently obtains state-of-the-art transfer ability on a variety of downstream tasks, such as image classification, object detection, and semantic segmentation.
\end{abstract}


\section{Introduction}
\label{sec:intro}

\IEEEPARstart{S}{elf-supervised} learning in computer vision (CV) \cite{zhang2016colorful,noroozi2016unsupervised,pathak2016context,komodakis2018unsupervised,wu2018unsupervised,hjelm2018learning,oord2018representation,he2019momentum,chen2020simple,chen2020improved,chen2020big,grill2020bootstrap,caron2020unsupervised,chen2020generative,jing2020self,dosovitskiy2021an,li2021mst,zhou2021ibot,bao2021beit,xie2021simmim,he2021masked,ye2022dynamic,dai2022unsupervised,huang2022learning,lin2022dual,huang2022survey,wang2022contrastive,qing2023self,wang2023caco,yang2023online} is widely used to learn general representations from large-scale unlabeled images without human annotation. 
Generally, massive pretext tasks \cite{hjelm2018learning,oord2018representation,he2019momentum,chen2020simple,chen2020improved,chen2020big,grill2020bootstrap,caron2020unsupervised,chen2020generative,dosovitskiy2021an,li2021mst,zhou2021ibot,bao2021beit,xie2021simmim,he2021masked} try to learn visual representations with self-supervision. 
Among them, inspired by the success of masked language modeling (MLM) \cite{devlin2018bert}, Masked image modeling (MIM) is used for pre-training in CV and has shown a preponderant advantage in performance. 
MIM masks patches of an image with a certain random mask ratio and predicts the original representations of the masked patches based on the remaining visible ones.
Commonly, the random mask ratio of MIM is much higher than that of MLM due to the difference in the information density of image and language data \cite{he2021masked}. For example, MAE \cite{he2021masked} adopts $75\%$  mask ratio while BERT \cite{devlin2018bert} uses $15\%$ one. However, we observe that the high random mask ratio brings two serious problems: 1) the data utilization of images is inadequate in each iteration, resulting in prolonged pre-training, and 2) the predictions of the pre-trained model exhibit the high inconsistency, bringing unreliable generations.  

Indeed, MAE exploits $25\%$ of the whole image in each iteration to train the model for $1600$ epochs, while BERT uses $85\%$ of the text corpus to train for only $40$ epochs. The reason of this phenomenon is not only the difference between images and language data, but also the difference between the MLM and MIM methods in the utilization of data. 
In terms of self-supervised learning in CV, the pre-training efficiency of MIM is still lower than that of contrastive learning. For example, MoCo v3 \cite{chen2021an} commonly pre-trains the model for $300$ epochs. The discrepancy in pre-training efficiency between contrastive learning and MIM stems from the fact that contrastive learning leverages the whole image for model training, thereby achieving a data utilization rate of $100\%$. From the above descriptions, it can be concluded that the high mask ratio of MIM leads to a low data utilization rate, bringing prolonged pre-training.

\begin{figure}[t]
    \centering
    \includegraphics[width=1.0\linewidth]{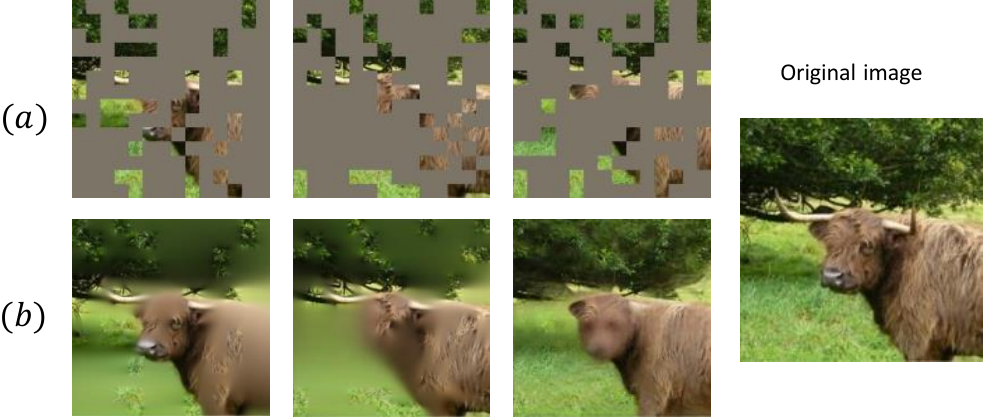}
    \caption{
        \textbf{Different reconstruction results of MAE \cite{he2021masked} correspond to different mask seeds.} (a) Different combinations sampled by different mask seeds. (b) Reconstruction results of MAE. For the three reconstructions of (b), only the first represents a normal cattle, and the third even reconstructs a dog. The semantics of these reconstructions by MAE are inconsistent.
    }
    \label{fig:figure1}
\end{figure}

Another problem of MIM is that the high mask ratio will introduce less reliable features. As shown in Figure~\ref{fig:figure1}, MAE generates different reconstruction results corresponding to different mask seeds, and the semantics of different results are inconsistent. That is, different combinations of random visible patches sampled from original images generate inconsistent predictions for identical positions. This inconsistency in generation is a contributing factor to unreliable results. Simultaneously, as a generation algorithm, if the generated results of MAE exhibit inconsistency and unreliability, then it does not acquire reliable and consistent representations, thereby undermining its transfer ability in various downstream tasks\cite{deng2009imagenet,he2016deep,ren2015faster,he2017mask,li2021benchmarking,chen2023mitigating,li2022transfering}. 

\begin{figure*}[ht]
    \centering
    \includegraphics[width=1.0\linewidth]{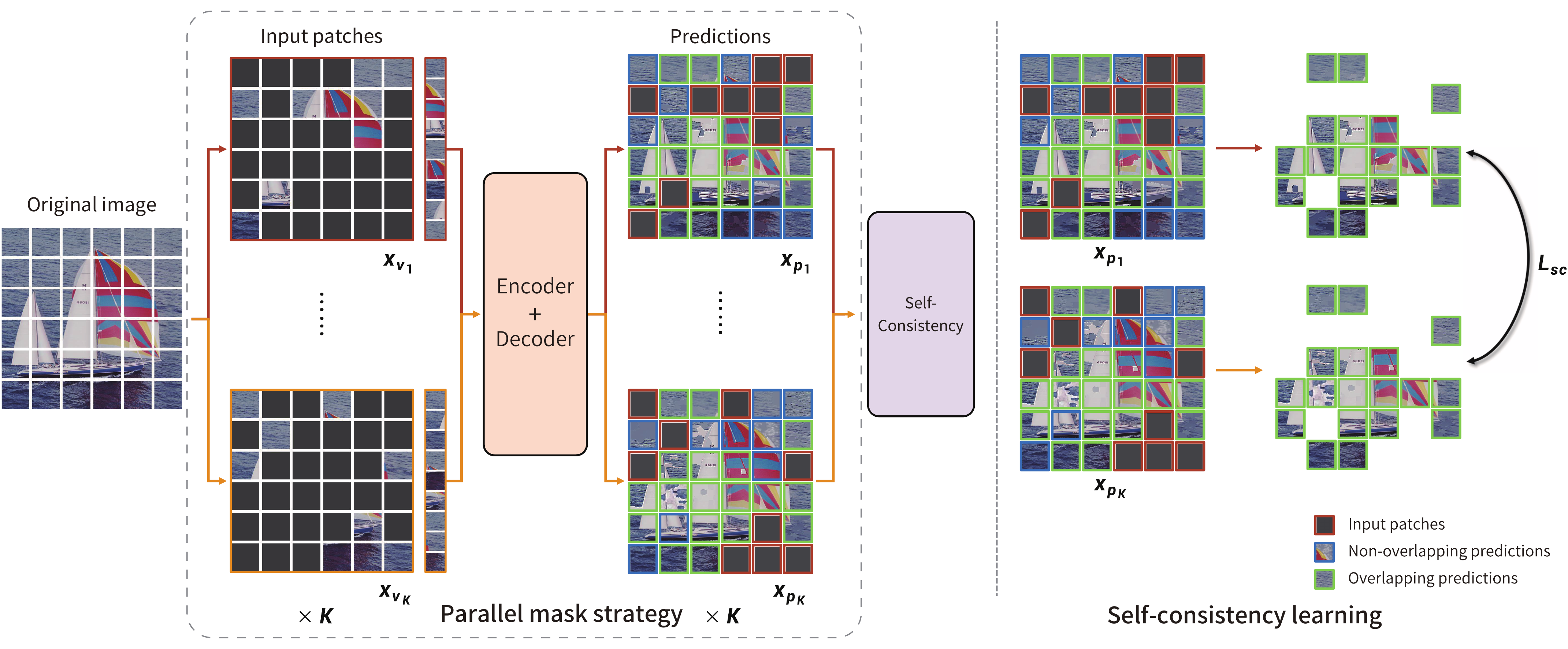}
    \caption{
        \textbf{Illustration of our EMAE.} The whole image is first divided into $N$ patches, then the parallel strategy divides $N$ patches into
        $K$ non-overlapping parts $\mathbf{x}_{v_1}$, $...$, $\mathbf{x}_{v_K}$ of the same size. Each part consists of $N/K$ random non-overlapping visible image patches. Then, each part is fed into the encoder-decoder architecture and performs the MIM task to generate $\mathbf{x}_{m_1}$, $...$, and $\mathbf{x}_{m_K}$. Furthermore, self-consistency learning guides the overlapping predictions in identical positions to be pulled together. Here, we take $\mathbf{x}_{p_1}$ and $\mathbf{x}_{p_K}$ as examples.
    }
    \label{fig:pipeline}
\end{figure*}

Is it possible to reduce the random mask ratio to increase the pre-training efficiency and improve the consistency? The answer is negative. In fact, the prior work \cite{he2021masked} already showed that reducing the mask ratio weakens the transfer ability of the pre-trained model for various visual downstream tasks. Therefore, it is important to address the above two problems posed by a high mask ratio while keeping the high mask ratio unchanged. In this paper, we propose efficient masked autoencoders with self-consistency (EMAE), which aims to enhance the pre-training efficiency and increase the certainty and consistency of MIM under the high mask ratio. Concretely, as illustrated in Figure~\ref{fig:pipeline}, a parallel mask strategy is proposed to divide the image into $K$ non-overlapping parts of the same size. Then, each part is a set that contains random non-overlapping visible image patches. Also, all parts have the same mask ratio and are treated in parallel. This way, the data utilization of images has been significantly enhanced, achieving a utilization rate of $100\%$. Meanwhile, the self-consistency learning is introduced to encourage the model to output consistent predictions in the identical positions of different parts, which increases the certainty and consistency of MIM. In addition, although the parallel mask strategy increases the computation amount of a single iteration, it reduces the overall number of training rounds and improves the pre-training efficiency.

We validate our method on a variety of visual pre-training datasets and visual downstream tasks.
With only $13\%$ of pre-training time using NVIDIA A100 GPUs, our method outperforms MAE in a variety of visual downstream tasks and achieves state-of-the-art (SOTA) performance. Moreover, under the architecture of ViT-Large, EMAE achieves $58.1\%$ bbox mAP and $50.4\%$ mask mAP using ViTDet \cite{li2022exploring} on COCO \cite{lin2014microsoft} object detection and instance segmentation. More experiments can refer to Experiments~\ref{sec:experiments}. 

Overall, we make the following contributions:

\begin{itemize}[leftmargin=0.2in]
    \item We observe that the high random mask ratio of MIM leads to two serious problems: prolonged pre-training and high inconsistency of the pre-trained model.
    
    \item We propose a parallel mask strategy to achieve $100\%$ of data utilization rate while maintaining the randomness, effectively improving the pre-training efficiency in MIM.
    \item Self-consistency learning is introduced into MIM and generates more reliable representations.
    \item Massive experiments demonstrate the effectiveness and stronger generalization ability of our method. In particular, EMAE outperforms previous SOTA methods in various downstream tasks, such as linear probing, fine-tuning classification, object detection, instance segmentation, and semantic segmentation.

\end{itemize}

\section{Related Works}
\label{sec:related_work}

Self-supervised learning \cite{zhang2016colorful,noroozi2016unsupervised,pathak2016context,komodakis2018unsupervised,wu2018unsupervised,hjelm2018learning,oord2018representation,he2019momentum,chen2020simple,chen2020improved,chen2020big,grill2020bootstrap,caron2020unsupervised,chen2020generative,jing2020self,dosovitskiy2021an,li2021mst,zhou2021ibot,bao2021beit,xie2021simmim,he2021masked,ye2022dynamic,dai2022unsupervised,huang2022learning,lin2022dual,huang2022survey,wang2022contrastive,qing2023self,wang2023caco,yang2023online,assran2022masked,nguyen2023r,baevski2022data2vec,assran2023self} builds general representations by exploiting the internal priors or structures of pre-training data. These representations can be transferred to various visual downstream tasks and improve the performance of these tasks. Self-supervised learning in computer vision is based on various pre-text tasks. In these work, a milestone in self-supervised learning was the contrastive learning \cite{wu2018unsupervised,he2019momentum,chen2020simple,chen2020improved,chen2020big,grill2020bootstrap,caron2020unsupervised}. In addition, mask image modeling \cite{chen2020generative,dosovitskiy2021an,li2021mst,zhou2021ibot,bao2021beit,xie2021simmim,he2021masked} has became the focus in the research community.
\subsection{Contrastive learning}

In the domain of computer vision, contrastive learning/instance discrimination \cite{he2019momentum,chen2020simple,chen2020improved,chen2020big,grill2020bootstrap,caron2020unsupervised} has been popular for self-supervised visual representation learning. The basic principle of these methods is that different views of an image are still the same category. This approach has proven the excellent performance on extensive visual downstream tasks, such as image classification \cite{deng2009imagenet,he2016deep}, object detection \cite{ren2015faster,li2021benchmarking}, and semantic segmentation \cite{he2017mask}. The pioneering work \cite{wu2018unsupervised} proposes to use a memory bank to store the instance class representation vector. Then, MoCo \cite{he2019momentum} improves the training of instance discrimination methods by storing representations from a momentum encoder instead of the trained network. Meanwhile, SimCLR \cite{chen2020simple} shows that the memory bank can be entirely replaced with the elements from the same batch if the batch is large enough. Besides, BYOL \cite{grill2020bootstrap} proposes an asymmetric structure and directly bootstraps the representations by attracting the different features from the same instance. It also shows that contrastive learning without negative samples can also learn excellent visual representations. Moreover, MoCo v3 \cite{chen2021an} applies the practice of contrastive learning in convolutional neural networks to ViT \cite{dosovitskiy2021an} architecture. DINO \cite{caron2021emerging} utilizes knowledge distillation together with ViT in the contrastive learning framework. MSN \cite{assran2022masked} proposes a masking invariance strategy to augment data for self-supervised learning. However, much of their success has far been limited to single-centric-object pre-training data such as ImageNet \cite{deng2009imagenet} due to the prior of image semantic consistency \cite{liu2020self,li2022univip}, and cannot be extended to non-iconic datasets. To overcome the problem, UniVIP \cite{li2022univip} proposes a unified self-supervised framework to learn general visual representations from arbitrary images, whether they are from the single-centric-object or non-iconic dataset. It is worth noting that our work focuses solely on the masked image modeling and is unrelated to contrastive learning.

\subsection{Masked image modeling}
With the development of vision Transformer \cite{dosovitskiy2021an,han2022survey,chen2023devil,chen2022obj2seq} and the success of MLM paradigm in NLP \cite{devlin2018bert}, MIM becomes the focus of the research community. MST \cite{li2021mst} is the first to introduce MIM into the siamese structure and proposes an attention-guided mask strategy. Furthermore, iBOT \cite{zhou2021ibot} also adopts the siamese structure and acquires impressive performance. Besides, some researchers combine contrastive learning and masked image modeling \cite{huang2022contrastive,quan2022global,mishra2022simple,yi2022masked,lee2022exploring,zhang2022mask,tao2022siamese,zhangcontextual} to achieve better performance.
The above studies benefit from the assumption of image/global semantic consistency of contrastive learning \cite{li2022univip}. Additionally, the prior work \cite{chen2020generative} operates on sequences of pixels and predicts unknown pixels. ViT \cite{dosovitskiy2021an} also attempts to study masked patch prediction for self-supervised learning. Following ViT, SimMIM \cite{xie2021simmim} proposes to predict pixels. Moreover, BEiT \cite{bao2021beit} proposes to predict discrete tokens, and its performance depends on the performance of the pre-trained model VQVAE \cite{van2017neural}. It is noted that MAE \cite{he2021masked} proposes an asymmetric encoder-decoder architecture for the MIM task and shows excellent performance in a variety of visual downstream tasks. R-MAE \cite{nguyen2023r} explores the use of regions for reconstructive pre-training. Data2vec \cite{baevski2022data2vec} extends MIM to masked multi-modality modeling. 
However, the mask ratio of MIM is commonly high and this results in two serious problems, namely prolonged pre-training and inconsistency of the pre-trained model. To overcome these problems, we propose efficient masked autoencoders with self-consistency to improve the pre-training efficiency and consistency of MIM under the condition of a high mask ratio. Notably, I-JEPA \cite{assran2023self} utilizes an online tokenizer to improve the efficiency of MIM, and is orthogonal to our method.

Furthermore, as a representation learning method for images, MAE alone fails to yield satisfactory image generation results. To address this problem, DiffMAE~\cite{wei2023diffusion} combines MAE with diffusion methods~\cite{dhariwal2021diffusion,ho2020denoising}, significantly improving the generation quality while maintaining representation quality. MAGE~\cite{li2023mage} employs the tokenizer from VQGAN~\cite{esser2021taming} as the embedding for MAE and uses discrete tokens as prediction targets, similarly enhancing generation quality without compromising representation quality. Moreover, RCG~\cite{li2023self} further proposes a self-conditioned generation approach based on MAGE. Our approach is orthogonal to these methods and can be combined with them to enhance the quality of image generation.

\subsection{Autoregressive image encoding}
Inspired by the success of GPT \cite{radford2018improving} in NLP, image GPT \cite{chen2020generative} predicts the next pixels based on the previous pixels and is the first work to introduce GPT to self-supervised learning in CV. 
Compared with MIM pre-training for ViT, GPT-like autoregressive models \cite{chen2020generative,hua2022self} in CV can utilize the whole image to perform the self-supervised task. However, the kind of method does not show superior performance and has quadratic time/space complexity of self-attention. 

\section{Methods}
In this section, we propose an efficient masked autoencoder with self-consistency (EMAE) to overcome the limitations of the high mask ratio in MIM. The pipeline of our proposed method is shown in Figure~\ref{fig:pipeline}. EMAE mainly consists of two main components. The first component is the parallel mask strategy, which achieves $100\%$ of the data utilization rate, improving the pre-training efficiency. Besides, EMAE introduces the self-consistency learning, which enforces consistency among different predictions of the identical positions. 

Here, we first give preliminaries about MAE in Section \ref{sec:preliminary}, which is adopted as our baseline of MIM. Then, we introduce the design of our proposed EMAE, a simple method that can greatly improve the pre-training efficiency and obtain reliable representations. The principle of EMAE boosting the learning efficiency by the parallel mask strategy is in Section \ref{sec:pa_mask} and the consistency of predictions by self-consistency learning is in Section \ref{sec:consis}. Moreover, the pre-training loss function is presented in Section \ref{sec:loss}.
Finally, we discuss how the proposed approach will affect efficiency and performance in Section \ref{sec:discussion}.  

\subsection{Preliminaries about MAE}
\label{sec:preliminary}

MAE is a highly recognized MIM framework for self-supervised learning, which gradually incorporates some of the impressive practices in the area. Specifically, according to the pioneering work \cite{he2021masked}, given a natural image from an unlabeled dataset $\mathbf{X}$, we divide it into $N$ regular image patches, denoted as $\mathbf{x} \in \mathbf{R}^{N\times S}$ where $S$ denotes the patch size ($e.g.$ $16\times16$ in ViT \cite{dosovitskiy2021an}). Then, we let mask $\mathbf{m}$ = $(m_1,...,m_N)$ denote a binary vector of length $N$, where $m_a \in \{0,1\}$ and $a \in \{1,2,...,N\}$, representing the mask over the image
, and generating two complementary combinations of $\mathbf{x}$: masked patches $\mathbf{x}_{m}$ and the visible patches $\mathbf{x}_{v}$ are given with Eq(\ref{eq:vom_mask}),

\begin{equation}
    \mathbf{x}_{m} = \mathbf{x} \odot \mathbf{m} \in \mathbf{R}^{\eta \times S} , \mathbf{x}_{v} = \mathbf{x} \odot (1- \mathbf{m}) \in \mathbf{R}^{\kappa \times S},
    \label{eq:vom_mask}
\end{equation}  
 
where $\eta = N \times p$ and $\kappa = N \times (1-p)$. The generation of mask $\mathbf{m}$ follows a uniform distribution, named the random mask strategy.

The MAE model $h=g\circ f$ is an encoder-decoder architecture, we fed these visible patches into encoder $f(\cdot)$ ($e.g.$, ViT-Base), and obtain the latent features. Then, a decoder $g(\cdot)$ maps the latent feature back to the pixel space to reconstruct the complementary combination and obtain the prediction $\mathbf{x}_{p}$. In particular, MAE minimizes the mean squared error (MSE) between the reconstructed and masked image patches $\mathbf{x}_{m}$, and the loss function is shown as Eq(\ref{eq:mae}).

\begin{equation}
    \mathcal{L}_{\text{MAE}}(\mathbf{x}) = \mathcal{L}(\mathbf{x}_{v},\mathbf{x}_{m}) = \mathop{\EE} ||g(f(\mathbf{x}_{v})) - \mathbf{x}_{m}||^{2},
    \label{eq:mae}
\end{equation}

\subsection{Parallel mask strategy}
\label{sec:pa_mask}

As described in Eq(\ref{eq:vom_mask}), the random mask ratio is $p$, thus the data utilization rate is $1 - p$. That is, the higher the random mask ratio, the lower the data utilization rate. The low data utilization rate diminishes the efficiency of pre-training due to inadequate involvement of training data. For instance, using MAE and BERT as examples of MIM and MLM respectively. If both models are pre-trained with $M$ epochs, then the ratio of data utilization rates between MLM and MIM is $(\frac{17}{5})^M$. If contrastive learning is applied instead of MLM, the data utilization ratio increases to $(4)^M$ because contrastive learning exploits the whole image. Besides, the prior work \cite{he2021masked} demonstrated that a high mask ratio ($75\%$) enhances the generalization of the pre-trained model. 
Hence, it is essential to achieve $100\%$ of data utilization rate under a high mask ratio to optimize pre-training and increase pre-training efficiency.

\begin{algorithm}[t]
\caption{Pseudocode of mask generation of the parallel mask strategy in a PyTorch-like style.}
\label{alg:code}

\definecolor{codeblue}{rgb}{0.25,0.5,0.5}
\lstset{
  backgroundcolor=\color{white},
  basicstyle=\fontsize{7.2pt}{7.2pt}\ttfamily\selectfont,
  columns=fullflexible,
  breaklines=true,
  captionpos=b,
  commentstyle=\fontsize{7.2pt}{7.2pt}\color{codeblue},
  keywordstyle=\fontsize{7.2pt}{7.2pt},
}
\begin{lstlisting}[language=python]
# x: the input image
# K: the number of non-overlapping parts

# map an image into multiple image patches 
x = patchify(x) 
N, D = x.shape  # length, dim

tensor = rand(N)  # tensor in [0, 1]

# sort the tensor in ascending order
ids = argsort(tensor) 

# acquire the position of each element
ids_tensor = argsort(ids)

# divide the whole data into K parts
for i in range(1,K+1):

    # obtain the i-th visible patches 
    ids_i = ids[(i-1)*(N/K):i*(N/K)]
    x_v_i = gather(x, dim=0, index=ids_i)

    # obtain the i-th mask 
    m_i = ones(N)
    m_i[(i-1)*(N/K):i*(N/K)] = 0
    m_i = gather(m_i, dim=0, index=ids_tensor)
    
    # obtain the i-th masked patches 
    x_m_i = x[m_i].reshape(-1,D)
\end{lstlisting}
\fontsize{7.2pt}{0em}\selectfont \texttt{rand}: returns a tensor filled with random numbers from a uniform distribution on the interval [0,1]; 
\vspace{0.2em}

\fontsize{7.2pt}{0em}\selectfont \texttt{gather}: gathers values along an axis specified by dim;
\vspace{0.2em}

\fontsize{7.2pt}{0em}\selectfont \texttt{argsort}: returns the indices that sort a tensor along a given dimension in ascending order by value.

\end{algorithm}

Intuitively, sampling multiple times from the same image can improve data utilization to a certain extent. However, the random mask strategy of MAE essentially follows the principle of sampling with replacement \cite{loeve2017probability}. Once multiple samplings are performed on the same image, there is a high probability of sampling overlap. Especially, as the value of $p$ increases, the likelihood of overlap occurring also increases. Therefore, in the finite sampling times, the mask strategy of MAE leads to two problems: 1) cannot guarantee the whole data utilization ($100\%$ of data utilization rate) like the contrastive learning or supervised learning, and 2) the sampling times of each patch from the same image are commonly different.
To overcome the above problems, in this paper, we propose a parallel mask strategy to exploit the whole data under a high mask ratio for MIM pre-training and guarantee each patch can be equally trained under finite sampling times.  

Concretely, the whole image is first divided into $N$ image patches. Then, the tensor $\mathbf{t}$ with random values of length $N$ is generated, and each value of the tensor fits the uniform distribution on the interval $[0,1]$. The tensor is sorted in ascending order by value and the sorted indices $\mathbf{ids}$ are obtained as Eq(\ref{eq:ids_sort}), where $\texttt{s}(\cdot)$ returns the indices that sort a tensor in ascending order by value.

\begin{equation}
    \mathbf{ids} = \texttt{s}(\mathbf{t}),  
    \label{eq:ids_sort}
\end{equation}

Here, we divide the sorted indices of length $N$ equally into $K$ non-overlapping parts $\mathbf{ids}_{1}$, $\mathbf{ids}_{2}$, $...$, and $\mathbf{ids}_{K}$ as shown in Eq(\ref{eq:reserve}), where $i \in \{1,2,...,K\}$.

\begin{equation}
    \mathbf{ids}_{i} = \mathbf{ids}[(i-1) \times \frac{N}{K}:i \times \frac{N}{K}], 
    \label{eq:reserve}
\end{equation}

Therefore, $N$ image patches can be divided equally into $K$ non-overlapping parts $\mathbf{x}_{v_1}$, $\mathbf{x}_{v_2}$, $...$, and $\mathbf{x}_{v_K}$ by the indices $\mathbf{ids}_{i}$, as shown in Eq(\ref{eq:non-over}), where $\texttt{d}(\cdot)$ denotes drawing values from input $\mathbf{x}$ according to the specified indices $\mathbf{ids}_{i}$. 

\begin{equation}
    \mathbf{x}_{v_i} = \texttt{d}(\mathbf{x}, \mathbf{ids}_{i}),
    \label{eq:non-over}
\end{equation}

The mask $\mathbf{m}_i$ of any part is given with Eq(\ref{eq:mask}), where $\texttt{ms}(\cdot)$ obtains the mask from $\mathbf{t}$ according to $\mathbf{ids}_{i}$.

\begin{equation}
    \mathbf{m}_i = \texttt{ms}(\mathbf{t}, \mathbf{ids}_{i}),
    \label{eq:mask}
\end{equation}

Any part $\mathbf{x}_{v_i}$ has $N/K$ visible patches, and its corresponding complementary view $\mathbf{x}_{m_i}$ has $N-N/K$ masked patches, which is defined as Eq(\ref{eq:ms_patch}). Hence, the mask ratio of any part is $(N-N/K)/K=(K-1)/K$. More details are described in Algorithm \ref{alg:code}.

\begin{equation}
    \mathbf{x}_{m_i} = \mathbf{x}\odot\mathbf{m}_{i},
    \label{eq:ms_patch}
\end{equation}

From the above description, image patches are divided into $K$ non-overlapping parts according to the principle of sampling without replacement \cite{loeve2017probability}. 
When $K$ is set to be $4$, the mask ratio of each part is $75\%$ (the same as the mask ratio of MAE). Our design is theoretically \emph{ensure the utilization of whole data and each patch can participate in training an equal number of times when sampling times are finite, even with just one iteration}. This characteristic cannot be guaranteed by the sampling with replacement strategy, which is why \emph{our design has a higher data utilization rate compared to the pure random mask strategy of MAE}. The design is called the parallel mask strategy, which can theoretically maintain the randomness of the mask while ensuring complete and strictly disjoint input and is more robust and stable than the pure randomness of MAE's mask strategy.

According to our design in Figure \ref{fig:pipeline}, each part containing visible patches is fed as the input into the encoder-decoder architecture and performs the MIM task, and the loss function is defined as Eq(\ref{eq:whole}). 

\begin{equation}
    \mathcal{L}_{whole}(\mathbf{x}) = \mathop{\EE}\limits_{i\in [1,K]}\mathcal{L}(\mathbf{x}_{v_i},\mathbf{x}_{m_i}),
    \label{eq:whole}
\end{equation}

\subsection{Self-consistency learning}

\label{sec:consis}

The parallel mask strategy in Section \ref{sec:pa_mask} ensures the utilization of whole data, but it can not guarantee the reliability of the output results for each part. 
According to the pioneering work \cite{lecky1945self}, human intelligence is a self-consistency system, which helps to efficiently learn and correct mistakes. It is reasonable to believe that artificial models can also improve training efficiency and consistency when introducing the self-consistency mechanism. For the MIM pre-trained model, the unreliable generations stem from the model generating inconsistent predictions for identical positions when presented with different insufficient information inputs from the same input. Therefore, when the self-consistency principle is introduced to MIM, 
the predictions of identical positions from the pre-trained model are encouraged to be consistent under different input visible patches from the same image.

In terms of Section \ref{sec:pa_mask}, each part has $N/K$ of the whole image and generates $N-N/K$ predictions. The predictions of each part are $\mathbf{x}_{p_1}$, $\mathbf{x}_{p_2}$, $...$, and $\mathbf{x}_{p_K}$. Obviously, there is a certain ratio of overlap between the predictions for any two sets of parts, and the ratio is $(K-2)/(K-1)$. The overlapping position $\mathbf{s}_{ij}$ of any two sets of predictions $\mathbf{x}_{p_i}$, $\mathbf{x}_{p_j}$ can be obtained by the mask $\mathbf{m}_i$ and $\mathbf{m}_j$, where $i, j \in \{1,2,...,K\}$ and $i \neq j$. The definition of $\mathbf{s}_{ij}$ is denoted as Eq(\ref{eq:self-consistency}).

\begin{equation}
    {\mathbf{s}_{ij}} = {\mathbf{m}_i} \ \cap \ {\mathbf{m}_j}, 
    \label{eq:self-consistency}
\end{equation}

Consequently, self-consistency learning is proposed to guide the predictions of each overlapping position to keep consistent. As shown in Figure \ref{fig:pipeline}, self-consistency learning pulls together the overlapping predictions between any two sets $x_{pi}$ and $x_{pj}$, which minimizes the mean absolute error between the overlapping reconstructed results to increase consistency. The \emph{self-consistency loss} is defined as Eq(\ref{eq:self-loss}), where $\texttt{sg[$\cdot$]}$ stands for stop gradient. For each prediction of any part, it will calculate with that of other parts by $K-2$ times to ensure self-consistency.

\begin{equation}
    \resizebox{1.0\hsize}{!}{$\begin{aligned}
    \mathcal{L}_{sc}(\mathbf{x}_{v_i},\mathbf{x}_{v_j}) = \mathop{\EE}\left(||\texttt{sg}[\mathbf{x}_{p_i}] - \mathbf{x}_{p_j} || + || \mathbf{x}_{p_i} - \texttt{sg}[\mathbf{x}_{p_j}]||\right) \odot {\mathbf{s}_{ij}}, 
    \end{aligned}$}
    \label{eq:self-loss}
\end{equation}

Finally, the self-consistency loss of the image is calculated according to the Eq(\ref{eq:total-self}).

\begin{equation}
    \mathcal{L}_{consistency} (\mathbf{x}) = \mathop{\EE}\limits_{i\in [1,K],\ j\in [i+1,K]}  \mathcal{L}_{sc}(\mathbf{x}_{v_i},\mathbf{x}_{v_j}), 
    \label{eq:total-self}
\end{equation}

The behavior induced by the self-consistency loss can be observed in Figure \ref{fig:figure2}: the reconstructed images from different combinations end up matching closely each other.

\subsection{Objective function}
\label{sec:loss}

Our EMAE consists of the design of the parallel mask strategy and self-consistency learning. Here, we still adopt the pixel reconstruction loss of MAE in order to preserve the generative ability of images.
Thus, the final loss for our EMAE can be formulated as Eq(\ref{eq:total-self}), and each loss coefficient is set to be $1$ for equally weighted.

\begin{equation}
    \mathcal{L}_{total} (\mathbf{x}) = \mathcal{L}_{whole}(\mathbf{x}) + \mathcal{L}_{consistency} (\mathbf{x}), 
    \label{eq:total-self}
\end{equation}

\subsection{Discussion and analysis}
\label{sec:discussion}

In this section, we present some intuitive analysis about why EMAE can improve the pre-training efficiency and increase consistency, which will be further demonstrated with empirical results in Section \ref{sec:experiments}. The primary component that makes EMAE converge faster is the utilization of multiple non-overlapping parts, which efficiently exploits the whole image in the training stage. Thus, EMAE can get more supervision signals in each epoch compared to MAE and achieves promising performance with fewer epochs. Notably, according to the parallel mask strategy, the whole image is divided equally into $K$ non-overlapping parts. The design can ensure each patch in an image can be used to train the model, thus enhancing the utilization of whole data. Finally, we propose the self-consistency loss to decrease the uncertainty and inconsistency of MIM. Based on the design of the whole data utilization, the self-consistency mechanism further improves feature representations, thereby benefiting the performance. Owing to these merits, EMAE can achieve high pre-training efficiency and consistent representations, thus obtaining promising performance.

\begin{table*}[t]
\centering
\caption{\textbf{Comparison of MAE and EMAE in time consumption.} `Total Iteration' displays the ratio of total iterations for different methods while `Total FLOPs per iteration' refers to the ratio of total iterations in a single iteration for different methods. Besides, `Total times' represents the performance of various methods to achieve its best: MAE trains 2400 epochs while EMAE and the supervised train 300 epochs. The K of EMAE is set to 4, and each mask ratio is 75\%. `Total speedup' shows the ratio of total speedup for different methods on 64 NVIDIA A100 GPUs.
}
\label{tab:consumption}
\scalebox{1.1}{
\begin{tabular}{lccccc}
Encoder & Architecture & Total iterations & Total FLOPs per iteration & Total times (hours) &Total speedup\\
\shline
MAE & ViT-L  & 1$\times$ & 1$\times$   & 180.76  & 1$\times$\\
Supervised & ViT-L& 0.125$\times$  & $\sim$ 4$\times$  & 23.83 &\bf{7.58}$\times$ \\
EMAE & ViT-L  & 0.125$\times$ & $\sim$ 4$\times$   & 23.75 &\bf{7.61}$\times$ \\
\hline
\end{tabular}
}
\end{table*}

\textbf{Time consumption.} In this paper, we point out that the high mask ratio leads to prolonged pre-training, EMAE proposes the utilization of whole data to improve the pre-training efficiency. The more data brings more calculations in the same epoch and therefore may lead to extra time consumption. However, the parallel mask strategy divides the image into K equal parts. This design can utilize the characteristics of the GPU for parallel acceleration and makes full use of the memory.
Unlike prior work \cite{chen2022context, dong2022bootstrapped}, which can only be calculated serially, EMAE can avoid plenty of time consumption. 
Simultaneously, we conduct the experiment to compare the time consumption of MAE, supervised learning, and EMAE.
Among them, MAE is trained for $2400$ epochs, while supervised learning and EMAE are trained for $300$ epochs. The reason for this choice is that the performance of 300-epoch EMAE is comparable to that of 2400-epoch MAE, and it is common for supervised learning methods to be trained for $300$ epochs. The comparison results of time consumption are illustrated in Table \ref{tab:consumption}. It is noted that 300-epoch EMAE can achieve the performance of 2400-epoch MAE (the SOTA performance) in only $13\%$ of time consumption. Moreover, by consuming only the same or even less time than the supervised, EMAE can achieve SOTA performance. Hence, all of these results indicate that EMAE is an efficient algorithm and can save pre-training times.

\begin{figure}[t]
    \centering
    \includegraphics[width=1.0\linewidth]{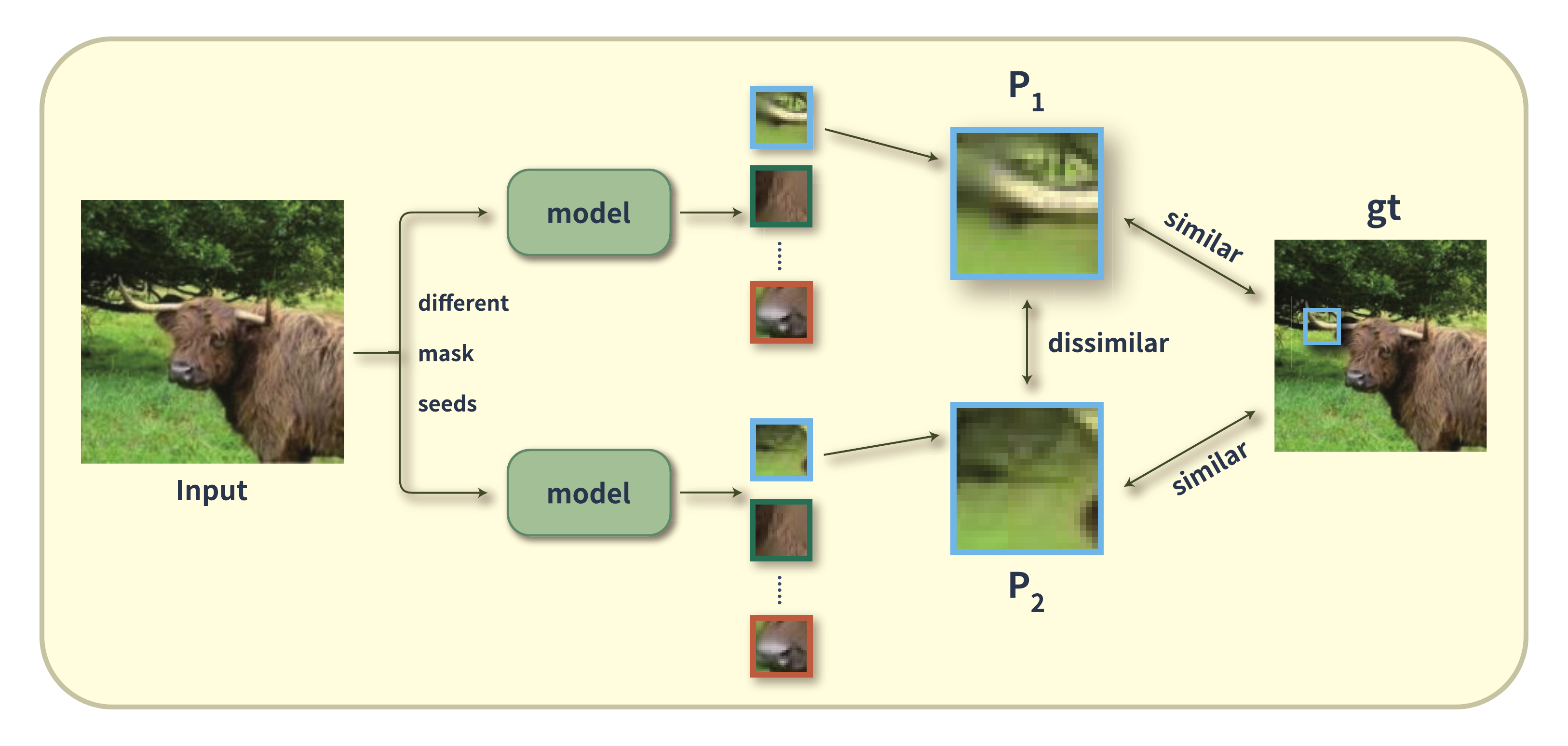}
    \caption{
        \textbf{Different predictions of identical positions of MIM pre-trained model using different mask seeds.} P$_{1}$ and P$_{2}$ are the predictions of identical positions, and gt is the ground truth. Both P$_{1}$ and P$_{2}$ are similar to gt, however, the semantics between P$_{1}$ and P$_{2}$ are dissimilar.
    }
    \label{fig:figure3}
\end{figure}

\textbf{How and why does self-consistency learning work?} The high mask ratio of MIM would introduce insufficient input information, making the learning task more challenging and resulting in the plateauing of loss reduction during the later stages of model pre-training. Consequently, the pre-trained model may generate semantic-inconsistent predictions even in identical positions. This unstable pre-training paradigm disrupts the learning of visual representations and undermines the transfer ability in various visual downstream tasks. As illustrated in Figure~\ref{fig:figure3}, two different parts of input using different mask seeds yield two dissimilar predictions even if they are all trained to align with the original image patch. Therefore, the predictions of identical positions require further constraint, in this paper, the constraint is self-consistency learning. Self-consistency learning encourages the predictions of identical positions to be consistent, thereby obtaining more reliable and consistent generations.
Finally, as shown in Figure~\ref{fig:figure2} in Section \ref{sec:experiments}, the stable reconstructed images further show the effectiveness of the self-consistency learning we proposed.

\section{Experiments}
\label{sec:experiments}

\subsection{Datasets and evaluation metrics}
\textbf{Pre-training datasets.} 
In the pre-training stage, we select three popular visual pre-training datasets, ImageNet-$1$K \cite{deng2009imagenet}, COCO \cite{lin2014microsoft}, and OpenImages \cite{kuznetsova2020open} to investigate the generality and transfer ability of the proposed method. First, ImageNet-$1$K dataset is a large-scale image classification dataset and its \texttt{train} set consists of $\sim$$1.28$ million training images with $1000$ categories. We conduct extensive experiments in the \texttt{train} set of ImageNet. Besides, COCO dataset is a non-iconic image dataset, the images in COCO contain natural and diverse scenes and have a scene-centric nature. Its \texttt{train2017} set contains about $118k$ images, and we conduct experiments in the set. Finally, OpenImages dataset is also non-iconic and contains a larger amount of data and categories compared to the COCO dataset. Its train set has $\sim$$1.7$ million images, and we also conduct experiments in the set. ImageNet is the default dataset.

\textbf{Evaluation datasets.} 
For validating the quality of pre-trained models, EMAE is evaluated by the linear probing and fine-tuning classification task on the ImageNet-$1$K dataset, which is the popular image classification dataset. Additionally, for computer vision, object detection and segmentation are dense prediction tasks since training images of these tasks contain massive objects inside each image. The validation of dense prediction tasks can better reflect the semantics capacity of pre-trained models. Therefore, we conduct extensive experiments on COCO \cite{lin2014microsoft} and ADE20k \cite{zhou2019semantic} datasets to verify the generalization and transfer ability of EMAE. COCO is a popular object detection and instance segmentation dataset, which \texttt{train2017} contains about $118k$ images and validate on the \texttt{val2017} contains $5k$ images. The challenging dataset contains human annotations for $80$ classes. Furthermore, ADE20K is also a challenging semantic segmentation dataset, and it contains about $20k$ training images and $2k$ validation images with $150$ categories.

\begin{table*}[t]
\tabcolsep=0.59cm
\centering
\caption{\textbf{Comparison of SOTA self-supervised learning methods.} 
EMAE is pre-trained on ImageNet \texttt{train} set and achieves state-of-the-art performance than previous masked image modeling methods. For evaluation, we test the performance of the pre-trained models under two supervised training settings: 1) linear probing, and 2) end-to-end fine-tuning. We report top-1 accuracy on the ImageNet \texttt{val} set.}
\small
\begin{tabular}{lccll}
Method & Architecture  & Pre-train epochs & Linear Probing Top-1 Acc & Fine-tuning Top-1 Acc  \\
\hline
\multicolumn{5}{l}{\emph{Supervised learning on ImageNet:}} \\
\gray{scratch \cite{chen2022context}} & \gray{ViT-S} &\gray{300}& \gray{-} & \gray{79.9\%} \\
\gray{scratch \cite{he2021masked}} & \gray{ViT-B} &\gray{300}& \gray{-} & \gray{82.3\%} \\
\gray{scratch \cite{he2021masked}} & \gray{ViT-L} &\gray{300}& \gray{-} & \gray{82.6\%} \\
\hline
\multicolumn{5}{l}{\emph{Contrastive learning:}} \\
MoCo v3~\cite{chen2021an} & ViT-B & 300 & 76.2\%& 83.2\% \\
DINO~\cite{caron2021emerging} & ViT-B & 400 & {\bf 78.2\%}& {\bf 83.4\%} \\
MSN~\cite{assran2022masked} & ViT-B & 600 & - & {\bf 83.4\%} \\
\hline
\multicolumn{5}{l}{\emph{Masked image modeling + contrastive learning:}} \\
MST~\cite{li2021mst} & ViT-S & 100 & 75.0\%& -\\
AttMask~\cite{kakogeorgiou2022hide}		& ViT-S & 100  & 76.1\%&81.3\% \\
iBOT~\cite{zhou2021ibot} & ViT-S & 100  & 74.4\%& 81.1\%\\
iBOT~\cite{zhou2021ibot} & ViT-S & 3,200  & 77.9\%& 82.3\%\\
iBOT~\cite{zhou2021ibot} & ViT-B & 1,600 & 79.5\%& 84.0\%\\
iBOT~\cite{zhou2021ibot} & ViT-L & 1,200 & \bf 81.0\%&\bf 84.8\%\\ 
\hline
\multicolumn{5}{l}{\emph{Masked image modeling:}} \\
CAE~\cite{chen2022context} & ViT-S  & 300  & 50.8\%& 81.8\% \\
CAE~\cite{chen2022context} & ViT-B  & 800  & 68.3\%& 83.6\% \\  
BEiT~\cite{bao2021beit} & ViT-B  & 800 & 56.7\%& 83.2\% \\
BEiT~\cite{bao2021beit} & ViT-L  & 800 & 73.5\%& 85.2\% \\
SimMIM~\cite{xie2021simmim} & ViT-B  & 800  & 56.7\%& 83.8\% \\
data2vec~\cite{baevski2022data2vec} & ViT-B  & 800& - & 84.2\% \\
data2vec~\cite{baevski2022data2vec}& ViT-L  & 1600& 77.3\% & 86.6\% \\
I-JEPA~\cite{assran2023self}& ViT-B  & 600& 72.9\% & - \\
I-JEPA~\cite{assran2023self}& ViT-L  & 600& \textbf{77.5}\% & - \\
MAE~\cite{he2021masked} & ViT-B  & 300& 61.5\% & 82.9\% \\
MAE~\cite{he2021masked} & ViT-B  & 800& 64.4\% & 83.4\% \\
MAE~\cite{he2021masked} & ViT-B  & 1600& 67.8\% & 83.6\% \\
MAE~\cite{he2021masked} & ViT-B  & 2400& 68.2\% & 83.8\% \\
MAE~\cite{he2021masked} & ViT-L  & 1600& 75.6\% & 85.9\% \\
MAE~\cite{he2021masked} & ViT-H  & 1600& 76.6\% & \textbf{86.9\%} \\
\hline
EMAE & ViT-B & 300  &  68.2\% &  83.8\% \\
EMAE & ViT-B & 800  &  70.4\% & 84.0\% \\
\textbf{EMAE} & ViT-L & 800  & \cellcolor{DnCBG}{\bf 76.7\%} & \cellcolor{DnCBG}{\bf 86.3\%} \\
\hline
\end{tabular}
\label{tab:classification}
\end{table*}

\textbf{Evaluation metrics.} 
Following common practices of self-supervised learning, we mainly use the top-1 accuracy to evaluate the semantic capacity of the pre-trained model for linear probing and fine-tuning image classification tasks. Meanwhile, for further validating the dense visual tasks, we adopt the box and mask mean average precision to validate the performance of transfer learning in object detection and instance segmentation tasks. Finally, we adopt the mean intersection of union to verify the transfer ability of the semantic segmentation task, which is also a popular dense visual task.

\subsection{Implementation}
\label{sec:imple}

\textbf{Pre-training settings.} 
The training settings are the same as MAE \cite{he2021masked}, we adopt the encoder-decoder structure to perform the MIM task. Our method is general for ViT backbones, while most experiments are conducted with ViT-Base, due to the limitation of computation resources. Specifically, we partition the image of $224\times 224$ into $14\times 14$ patches with the patch size being $16\times 16$. The $K$ is set to be $4$ by default. The batch size is set to be $4096$. Meanwhile, the weight decay, $\beta_1$ and $\beta_2$ for AdamW \cite{loshchilov2017decoupled} is set to be $0.05$, $0.9$ and $0.95$ respectively. We use a cosine learning rate strategy \cite{loshchilov2016sgdr} with $40$ epochs warmup \cite{goyal2017accurate}. The base learning rate is set to be $\textit{base\_lr} = 1.5e^{-4}$. The difference in pre-training setups across different datasets is solely reflected in the number of epochs: ImageNet is pre-trained for $800$ epochs, COCO dataset is trained for $4000$ epochs according to R-MAE \cite{nguyen2023r}, and OpenImages dataset is pre-trained for $800$ epochs following long-sequence MAE \cite{hu2022exploring}. Please refer to the Appendix for more details.

\textbf{Linear probing settings.}
Following common practice, we evaluate the representation quality by linear probing. After self-supervised pre-training, we remove the decoder and train a supervised linear classifier on frozen features. All training settings are the same as MAE~\cite{he2021masked}. As shown in the Appendix, the evaluated configuration of linear probing is listed.

\textbf{End-to-end finetuning settings.}
According to MAE~\cite{he2021masked}, the hyper-parameters of the end-to-end fine-tuning classification task are shown in the Appendix.

\textbf{Object detection and instance segmentation settings.}
For benchmarking setting, we train all models with the same simple formula: large-scale jitter~\cite{ghiasi2021simple} ($1024 \times 1024$ resolution, scale range $[0.1, 2.0]$, AdamW $(\beta_1, \beta_2 = 0.9, 0.999)$ with half-period cosine learning rate decay, linear warmup $0.25$ epochs, and drop path regularization ($e.g.$ $0.1$ for ViT-Base). All models are trained with 100 epochs and the batch size is set to be $64$. Also, the learning rate is $8e-5$, and the weight decay is $0.1$. More details are described in \cite{li2021benchmarking}, and we strictly follow the hyper-parameters of MAE. For the setting of ViTDet, the input size is $1024\times1024$, augmented during training by large-scale jitter with a scale range of $[0.1, 2.0]$. We use AdamW $(\beta_1, \beta_2{=}0.9, 0.999)$ with step-wise learning rate decay. We use linear learning rate warm-up with batch size 64 for 250 iterations. More details are described in \cite{li2022exploring}, and we fully follow it to conduct experiments.

\textbf{Semantic segmentation settings.}
We adopt UperNet framework \cite{xiao2018unified} in our experiments. We use AdamW as the optimizer. The batch size is $16$ and the layer-wise decay rate is $0.65$. The input resolution is $512 \times 512$. We finetune end-to-end for $100$ epochs. In fact, all experimental settings are consistent with MAE, and mmsegmentation \cite{mmseg2020} also open-sources the MAE segmentation code to help researchers to quickly verify.

\textbf{Data Augmentor settings.} According to prior work \cite{xu2022masked}, MAE can be utilized as a data augmentor. To validate the performance of MAE and EMAE in data augmentation, we employed 800-epoch EMAE and 1600-epoch MAE as data augmentors for image classification. This experiment utilizes ResNet50 \cite{he2016deep} as the backbone, conducts 90 epochs on the ImageNet dataset, and reports Top-1 accuracy on the validation set. Each image was subjected to the random mask setting rather than the attention-guided mask strategy \cite{li2021mst} to better showcase the generation quality of MAE and EMAE. Other hyperparameters follow the training recipes in \cite{he2016deep,li2023freconv}.

\subsection{Image classification on ImageNet-1K}

\textbf{Assessment setting.}
For a fair comparison, we strictly follow the hyperparameters of MAE \cite{he2021masked} in our image classification experiments to evaluate the ImageNet pre-trained models. We evaluate the performance of the pre-trained encoder under two supervised training settings: 1) linear probing (LP), and 2) end-to-end fine-tuning (FT).
For the linear probing, all parameters of the pre-trained encoder are frozen while only the last classification layer is trained. Also, for the end-to-end fine-tuning, the pre-trained encoder is fine-tuned with the classification head together.

\begin{table*}[t]
    \centering
    \caption{\textbf{Results of object detection and instance segmentation on COCO using Mask R-CNN \cite{li2021benchmarking,he2021masked}.} The architecture of various methods adopts ViT-B. We adopt Mask R-CNN \cite{he2017mask} with FPN \cite{lin2017feature}, and report the bounding box AP and mask AP on COCO \texttt{val2017}. EMAE outperforms the previous SOTA self-supervised learning method. $\dagger$: These results from \cite{chen2022context}.
    }
    \label{tab:coco}
\scalebox{1.22}{
\begin{tabular}{l |c| c|   l| l |l |  l | l | l}
        \hline
        \multirow{2}{*}{Method} & 
        \multirow{2}{*}{\begin{tabular}[c]{c}Pre-train\\ epochs\end{tabular}} & 
        \multirow{2}{*}{\begin{tabular}[c]{c}Pre-train\\ data\end{tabular}} & 
        \multicolumn{3}{c}{Object detection}& 
        \multicolumn{3}{c}{Instance segmentation}\\ 
        \cline{4-9}
          &    &   &{ $\text{AP}^{b}$} & {$\text{AP}^{b}_{50}$} & {$\text{AP}^{b}_{75}$} & 
        {$\text{AP}^{m}$} & 
        {$\text{AP}^{m}_{50}$} & 
        {$\text{AP}^{m}_{75}$}\\ 
        \hline
        Supervised \cite{chen2022context}  & 300 &ImageNet-$1$K  &  47.9\%  &  -  &  -  &  42.9\%  &  -  &  -  \\
        MoCo v3 \cite{chen2021an} & 600 &ImageNet-$1$K &  47.9\%  & - &  -  &  42.7\%  &  -  &  -  \\
        BEiT \cite{bao2021beit}& 800   &ImageNet-$1$K + DALLE &  49.8\%  &  -  &  - &  44.4\%  &  -  &  - \\
        CAE$\dagger$ \cite{chen2022context} & 1600  &ImageNet-$1$K &   50.0\%  &  70.9\%  &  54.8\%  &  44.0\% &  67.9\%  &  47.6\%  \\
        MAE \cite{he2021masked}& 1600    &ImageNet-$1$K &   50.4\%  &  70.8\%  &  55.7\%  &  44.9\%  &  68.3\%  &  48.9\%  \\
        \hline
        EMAE & 300  &ImageNet-$1$K &   50.6\%  &  70.9\%  &  56.1\%  &  45.0\% &  68.6\%  &  49.3\% \\
		\textbf{EMAE} & 800  &ImageNet-$1$K &   \textbf{51.4\%}  &  \textbf{72.2\%}  &  \textbf{56.5\%}  &  \textbf{45.7\%} &  \textbf{69.4\%}  &  \textbf{49.8\%}  \\
        \hline   
    \end{tabular} }
\end{table*}

\textbf{Classification results.} As shown in Tab.~\ref{tab:classification}, we surprisingly find that our method using ViT-Base can surpass the MAE by around $6.0\% \sim 6.7\%$ with the same pre-training epochs (300 and 800 epochs) in linear probing. The 300-epoch classification results are on par with that of the 2400-epoch MAE, while achieving this level of performance with EMAE only requires $13\%$ of the time and about $50\%$ of calculations consumed by MAE. The phenomenon indicates that our method can significantly improve the training efficiency of MIM. Extra training (800 epochs) further improves the linear result to $70.4\%$ and finetuning result to $84.0\%$, and achieves SOTA performance. Moreover, it is noted that our 800-epoch classification results using ViT-Large are comparable with that of 1600-epoch MAE using ViT-Huge. As a MIM-based method, EMAE surpasses previous SOTA MIM-based methods. The linear results of EMAE are slightly inferior to contrastive-based methods due to the contrastive-based method has the assumption of image semantic consistency \cite{liu2020self,li2022univip}, and the assumption is consistent with the prior of the linear probing task. In the following sections, we comprehensively evaluate the transfer ability of our EMAE through conducting more experiments in different datasets and tasks.

\subsection{Object detection and instance segmentation}
To further validate the learned visual representation of our EMAE pre-trained on ImageNet, we perform fine-tuning on the COCO \cite{lin2014microsoft} object detection and instance segmentation. We choose the Mask R-CNN \cite{he2017mask} framework. Concretely, we adopt FPNs \cite{lin2017feature} to scale the feature map into different sizes as introduced in \cite{li2021benchmarking}. By fully following the strategy of previous \cite{he2021masked,li2021benchmarking}, we conduct these experiments on COCO. The results are reported in Table \ref{tab:coco} in terms of box AP ($AP^{b}$) for detection and mask AP ($AP^{m}$) for segmentation. The table shows the performance of the learned representation by different self-supervised methods and supervised training. It can be observed that the results of 300-epoch EMAE are better than that of 1600-epoch MAE. Furthermore, our 800-epoch EMAE achieves the best results with $51.4 \%$ $AP^{b}$ and $45.7 \%$ $AP^{m}$, and surpasses 1600-epoch MAE by $1.0$ and $0.8$ points, individually. 

Besides, we also conduct experiments on the SOTA ViT-Based detection framework ViTDet \cite{li2022exploring} implemented in Detectron2 \cite{wu2019detectron2} to verify the transfer ability of MAE. For a fair comparison, all of these experiments strictly adopt the training settings of ViTDet. In Table \ref{tab:cascade_coco}, it can be observed that the object detection and instance segmentation results of 300-epoch EMAE all outperform that of 1600-epoch MAE in various metrics. Meanwhile, our 800-epoch EMAE outperforms 1600-epoch MAE $1.3$ and $1.2$ points on $AP^{b}$ and $AP^{m}$. Also, the results of ViTDet with Cascade Mask RCNN reach $54.8 \%$ $AP^{b}$ and $47.6 \%$ $AP^{m}$, and surpass MAE by $0.8$ and $0.9$ points, respectively. 

Moreover, we further perform object detection and instance segmentation experiments using ViTDet under ViT-Large architecture. The results are shown in Table~\ref{tab:coco_large}. The results of 800-epoch EMAE show consistent improvements in the large network structure, whether it is loaded in ViTDet with/without Cascade Mask RCNN.

These experiments show that our EMAE can be applicable to arbitrary architecture and detection framework to boost performance without additional training costs and efforts.

\begin{table}[t]
\centering
\caption{\small{
\textbf{Results of object detection and instance segmentation fine-tuned on COCO using ViTDet \cite{li2022exploring}}. For a fair comparison, the architecture of various methods adopts ViT-B. ViTDet \cite{li2022exploring} is adopted as the detection framework, and our EMAE achieves impressive performance and outperforms the previous SOTA self-supervised learning method MAE. 
}
}
\label{tab:cascade_coco}
\scalebox{0.97}{
\begin{tabular}{l|c|c|l|l}
    \hline
    \multirow{2}{*}{Method} & \multirow{2}{*}{\begin{tabular}[c]{c}Pre-train\\ data\end{tabular}} & \multirow{2}{*}{\begin{tabular}[c]{c}Pre-train\\ epochs\end{tabular}}  & \multirow{2}{*}{AP$^{b}$} & \multirow{2}{*}{AP$^{m}$} \\ &&&&\\ 
    \hline
    \gray{Rand Init}\cite{li2022exploring} & -  & - &  \gray{48.1\%}  & \gray{42.6\%}  \\
    Supervised\cite{li2022exploring} & ImageNet-$1$K & 300 & 47.6\% & 42.4\%   \\ 
    Supervised\cite{li2022exploring} & ImageNet-$21$K & 300 & 47.8\% & 42.6\%   \\ 
    \hline
    MAE\cite{he2019momentum}   & ImageNet-$1$K  & 1600  & 51.2\%  &45.5\% \\ 
    MAE, \emph{our impl.}   & ImageNet-$1$K  & 1600  & 51.6\%  &45.9\% \\
    R-MAE \cite{nguyen2023r}    &ImageNet-$1$K& 1600 &   52.3\% &46.4\% \\
    EMAE   & ImageNet-$1$K  & 300  & 51.8\%  &46.1\% \\
    \textbf{EMAE}  & ImageNet-$1$K   & 800  &\bf{52.5\%} &\bf{46.7\%} \\
    \hline
    MAE + cascade  & ImageNet-$1$K  & 1600  &54.0\% &46.7\% \\ 
    EMAE + cascade  & ImageNet-$1$K  & 300  &54.3\%  &46.9\% \\
    \textbf{EMAE + cascade}  & ImageNet-$1$K  &  800  &\bf{54.8\%} &\bf{47.6\%} \\
    \hline
\end{tabular}
}
\end{table}

\begin{table}[t]
\centering
\caption{
\textbf{Results of object detection and instance segmentation finetuned on COCO using ViTDet \cite{li2022exploring}}. For a fair comparison, the architecture of various methods adopts ViT-L. ViTDet \cite{li2022exploring} is adopted as the detection framework, and our EMAE achieves impressive performance and outperforms the previous SOTA self-supervised learning method MAE. 
}
\label{tab:coco_large}
\scalebox{1.22}{
\begin{tabular}{l|c|l|l}
    \hline
    Method & Pre-train data  & AP$^{b}$ & AP$^{m}$ \\ 
    \hline
    \gray{Rand Init}\cite{li2022exploring} & -  & \gray{50.0\%}  & \gray{44.2\%}  \\
    Supervised\cite{li2022exploring} & ImageNet-$1$K  & 49.6\% &43.8\%   \\ 
    Supervised\cite{li2022exploring} & ImageNet-$21$K  & 50.6\% &44.8\%   \\ 
    \hline
    MAE\cite{he2019momentum}   & ImageNet-$1$K    & 54.6\%  &48.6\% \\ 
    \textbf{EMAE}  & ImageNet-$1$K     &\bf 55.6\%  &\bf 49.4\% \\
    \hline
    MAE + cascade  & ImageNet-$1$K    &57.6\%  &50.0\% \\ 
    \textbf{EMAE + cascade}  & ImageNet-$1$K     &\bf 58.1\% &\bf 50.4\% \\
    \hline
\end{tabular}
}
\end{table}

\begin{table}
    \centering
    \caption{
   \textbf{Results of semantic segmentation on ADE$20$K using UperNet \cite{xiao2018unified}}. For a fair comparison, the architecture of various methods adopts ViT-B. We report results measured by mean Intersection of Union (mIoU), and EMAE surpasses the previous self-supervised method and achieves the SOTA performance. $\dagger$ represents that the experimental setting of semantic segmentation follows the prior work \cite{chen2022context}.
    } 
    \label{tab:segmentation}
    \setlength{\tabcolsep}{5pt}
\renewcommand{\arraystretch}{1.15}
    \footnotesize
    \scalebox{1.06}{
        \begin{tabular}{l c c c }
        \hline
            \multirow{2}{*}{Method}   & \multirow{2}{*}{\begin{tabular}[c]{c}Pre-train\\ data\end{tabular}} & \multirow{2}{*}{\begin{tabular}[c]{c}Pre-train\\ epochs\end{tabular}}  & {ADE} 
             \\
            \cline{4-4}
             & & & mIoU  \\
        \hline
        Supervised~\cite{he2021masked}  & ImageNet-$1$K & 300 & 47.4\% \\
        SplitMask~\cite{el2021large}  & ADE$20$K & 21000 & 45.7\%   \\
        MoCo v3 \cite{chen2021an}  & ImageNet-$1$K & 600 & 47.3\%   \\
        BEiT~\cite{he2021masked}  & ImageNet-$1$K+DALLE & 800 & 47.1\%\\
        PeCo~\cite{dong2021perceptual}  & ImageNet-$1$K & 300  & 46.7\%   \\
        CIM~\cite{fang2022corrupted}  & ImageNet-$1$K & 100  & 43.5\% \\
        R-MAE~\cite{nguyen2023r}    &ImageNet-$1$K& 1600 &   47.6\% \\
        MAE~\cite{he2021masked}  & ImageNet-$1$K & 1600  & 48.1\%   \\
        CAE$\dagger$~\cite{chen2022context}  & ImageNet-$1$K & 800  & 49.7\%   \\
        \hline
        EMAE  & ImageNet-$1$K & 300  & 48.4\%   \\
		\textbf{EMAE}   & ImageNet-$1$K & 800 & 49.3\%   \\
  \textbf{EMAE}$\dagger$   & ImageNet-$1$K & 800 & \textbf{50.8\%}    \\
        \hline   
    \end{tabular} }
\end{table}

\subsection{Semantic segmentation}
We also evaluate our ImageNet pre-trained EMAE on another dense prediction task, semantic segmentation on the ADE20K \cite{zhou2019semantic} dataset. The mean Intersection of Union (mIoU) averaged over all semantic categories is reported as the evaluation metric. In particular, by fully following the training settings of MAE, we adopt the UperNet framework \cite{xiao2018unified} in our experiments and report the results in Table \ref{tab:segmentation}. We compare our method with supervised pre-training on ImageNet-1K as well as SOTA self-supervised methods. It can be observed that the performance of the proposed EMAE outperforms that of 1600-epoch MAE when EMAE is pre-trained for $300$ epochs. Also, 800-epoch EMAE achieves the highest $49.3\%$ mIoU and gets superior performance than all the other baselines, further validating the effectiveness of our framework. 

Besides, we noticed that the current methods based on MAE, like Ge$^2$-AE \cite{liu2023devil}, CAE \cite{chen2022context}, MVP \cite{wei2022mvp}, MILAN \cite{hou2022milan}, adopt different hyperparameters to downstream tasks for better performance. The above experiments strictly follow the setting of MAE, and the results already demonstrate the effectiveness of EMAE. Here, for a fair comparison, we follow the semantic segmentation settings of CAE \cite{chen2022context} for our pre-trained model, and the result with ViT-Base is $50.8\%$ mIoU on ADE20k dataset and achieves the SOTA performance. 

\begin{table}
    \centering
    \caption{ \textbf{The data augmentation results.} We report ImageNet-$1$K classification results. The architecture of various methods adopts ViT-L.
    } \label{tab:data_aug}
    \setlength{\tabcolsep}{5pt}
\renewcommand{\arraystretch}{1.15}
    \footnotesize
    \scalebox{1.25}{
        \begin{tabular}{l l c  l }
        \toprule
        Model  & Augmentor   & Mask ratio &Top-1   \\
        \midrule
     \multirow{5}*{ResNet50\cite{he2016deep}}  & - & - & 76.3\% \\
        \cline{3-4}
      &  \multirow{2}*{MAE~\cite{he2021masked} }& 40\%& 77.3\%~$\green{\pmb{\uparrow}}$ \\
        \cline{3-4}
      &  & 75\%& 75.9\%~$\red{\pmb{\downarrow}}$ \\
      &  \multirow{2}*{EMAE} & 40\% & 78.1\%~$\green{\pmb{\uparrow}}$ \\
        \cline{3-4}
     &    & 75\% & 77.2\%~$\green{\pmb{\uparrow}}$ \\
        \bottomrule     
    \end{tabular} }
\end{table}

\begin{table}[t]
    \centering
    \caption{ \textbf{The COCO pre-training results.} We report COCO detection, instance segmentation, and ADE20k semantic segmentation results. The architecture of various methods adopts ViT-B.
    } \label{tab:coco_pre-training}
    \setlength{\tabcolsep}{5pt}
\renewcommand{\arraystretch}{1.15}
    \footnotesize
    \scalebox{1.15}{
        \begin{tabular}{l c  c c c}
        \toprule
            Method   & Pre-train epochs & AP$^{b}$ &AP$^{m}$
            & mIoU  \\
        \midrule
        MAE~\cite{he2021masked} & 2000&49.9\% & 44.5\% & 45.2\% \\
        MAE~\cite{he2021masked} & 4000&50.1\% & 44.6\% & 45.9\% \\
        MAE~\cite{he2021masked} & 8000&50.1\% & 44.6\% & 46.5\% \\
        R-MAE~\cite{nguyen2023r} & 2000&49.7\% & 44.1\% & 46.0\% \\
        R-MAE~\cite{nguyen2023r} & 4000&50.6\% & 45.0\% & 46.8\% \\
        R-MAE~\cite{nguyen2023r} & 8000&50.8\% & 45.2\% & 47.0\% \\
        EMAE & 1000 & 50.0\% & 44.5\%&46.6\% \\
        EMAE & 2000 & 50.7\% & 45.1\% &47.2\% \\
        EMAE & 4000 & \textbf{51.0\%} & \textbf{45.3\%} & \textbf{47.7\%} \\
        \bottomrule     
    \end{tabular} }
\end{table}

\subsection{Data augmentor}
To showcase the generation quality of EMAE, we employ two different random mask ratios during data augmentation for ResNet image classification experiments: the one at $40\%$ and the other at $75\%$. The final experimental outcomes are presented in Table \ref{tab:data_aug}. Both MAE and EMAE could enhance the recognition accuracy of ResNet under the $40\%$ random mask ratio, indicating the viability of masked models as data augmenters. However, under the $75\%$ setting, the performance of ResNet trained on the data generated by MAE fell below the baseline, whereas the one generated by EMAE still improved ResNet's performance. This demonstrates that EMAE is able to generate more reliable and stable results than MAE does and indeed addresses the issue of inconsistent generation brought about by the high mask ratio.

\subsection{Pre-training on COCO dataset}
To demonstrate the adaptability of EMAE, we conduct pre-training experiments on COCO dataset. We assess the performance of pre-trained models on benchmark tasks according to R-MAE \cite{nguyen2023r}: object detection, instance segmentation, and semantic segmentation. The EMAE is primarily trained with $1000$, $2000$, and $4000$ epochs, respectively. For a fair comparison, we adopt ViTDet as the framework of object detection and instance segmentation according to R-MAE, and the results of different methods are shown in Table \ref{tab:coco_pre-training}. It can be observed that the performance of the 1000-epoch EMAE closely approximates that of the 8000-epoch MAE, showcasing the efficiency of our algorithm. Furthermore, the results from the 2000-epoch EMAE are comparable to the 8000-epoch R-MAE, while the 4000-epoch EMAE achieves SOTA performance. All these findings demonstrate that EMAE can acquire effective representations of non-iconic data.

\subsection{Pre-training on OpenImages dataset}
To further demonstrate the adaptability of EMAE, we conduct pre-training experiments on OpenImages dataset. We evaluate the performance of object detection, instance segmentation, and semantic segmentation tasks according to long-sequence MAE \cite{hu2022exploring}. The experimental settings of various downstream tasks follow the default ones. Long-sequence MAE is a method that pre-trains MAE by upscaling images from 224 resolution to 448 one. For a fair comparison, we do not introduce the results of long-sequence MAE in Table \ref{tab:openimages_pre-training}. OpenImages contains a larger amount of data and categories compared to the COCO dataset. This indicates that OpenImages is a dense object dataset with more complex scenes. In scenarios with a high mask ratio, a substantial number of objects are likely to be fully masked, rendering the task of mask reconstruction more challenging.  Nonetheless, even with such challenge, the results of 100-epoch EMAE still outperform that of 800-epoch MAE according to Table \ref{tab:openimages_pre-training}. This underscores the effectiveness of EMAE in achieving stable, reliable, and efficient training on dense object data. Furthermore, the performance of 800-epoch EMAE achieves $50.8\%$ $AP^{b}$ and $45.3\%$ $AP^{m}$ on COCO dataset and $49.3\%$ mIoU on ADE20k dataset. The results of pre-training EMAE on various datasets indicate its robustness to training data across diverse scenarios.

\begin{table}
    \centering
    \caption{ \textbf{The OpenImages pre-training results.} We report COCO detection, instance segmentation, and ADE20k semantic segmentation results. The architecture of various methods adopts ViT-B.
    } \label{tab:openimages_pre-training}
    \setlength{\tabcolsep}{5pt}
\renewcommand{\arraystretch}{1.15}
    \footnotesize
    \scalebox{1.15}{
        \begin{tabular}{l  c c c c}
        \toprule
            Method   & Pre-train epochs & AP$^{b}$ &AP$^{m}$
            & mIoU  \\
        \midrule
        MAE~\cite{he2021masked} & 800&49.9\% & 44.6\% & 47.8\% \\
        EMAE &  100 & 50.0\% & 44.8\% &48.0\% \\
        EMAE &  800 & \textbf{50.8\%} & \textbf{45.3\%} & \textbf{49.3\%} \\
        \bottomrule     
    \end{tabular} }
\end{table}

\subsection{Ablation studies}
\label{sec:ablation}
To better investigate the effectiveness of different components in our proposed EMAE, we conduct ablation studies on ImageNet-$1$K dataset. Linear probing is still an excellent evaluation method to quickly validate the learned representations, and MAE also selects the mask ratio according to the performance of linear probing. Hence, we adopt the results of linear probing as our benchmark for measuring effectiveness in ablation studies. For a fair comparison, the architecture of various
methods adopts ViT-Base.

\begin{table}[t]
\centering
\caption{\textbf{Ablations for EMAE: Effect of whole data utilization.} We report the results of linear evaluation on ImageNet.}
\label{tab:whole_data}
\scalebox{0.95}{
\begin{tabular}{l|c|c|c}
\shline
Method & Pre-train data  &  Pre-train epochs & Top-1  \\ \shline
(a) baseline & \multirow{4}{*}{ImageNet-$1$K} & \multirow{4}{*}{200} & 58.8\%  \\ \cline{0-0} \cline{4-4}
(b) pure random mask  & &   & 62.5\%  \\ \cline{0-0} \cline{4-4}
(c) parallel mask & &  & 63.4\%  \\ \cline{0-0} \cline{4-4}
(d) complementary mask  & &   & 46.7\%  \\ 
\shline
\end{tabular}}
\end{table}

\begin{table}[t]
\centering
\caption{\textbf{Ablations for EMAE: Effect of self-consistency learning.} We report the results of linear evaluation on ImageNet.}
\label{tab:self-con}
\scalebox{0.95}{
\begin{tabular}{l|c|c|c}
\hline
 \multirow{2}{*}{Method} & \multirow{2}{*}{\begin{tabular}[c]{c}Pre-train\\ data\end{tabular}}  &  \multirow{2}{*}{\begin{tabular}[c]{c}Pre-train\\ epochs\end{tabular}} & \multirow{2}{*}{Top-1}  \\ &&& \\ \shline
(a) pixel reconstruction & \multirow{5}{*}{ImageNet-$1$K} & \multirow{5}{*}{200} & 63.4\%  \\ \cline{0-0} \cline{4-4}
(b) pure self-consistency  & &   & 64.9\%  \\ \cline{0-0} \cline{4-4}
 \multirow{3}{*}{(c) \begin{tabular}[c]{c}pixel reconstruction\\ + \\self-consistency\end{tabular}}   & &  & \multirow{3}{*}{65.0\%}  \\ &&& \\ &&& \\ 
\hline
\end{tabular}}
\end{table}

\begin{table}
\centering
\caption{\textbf{Ablations for EMAE: Effect of the proposed method on different pre-training epochs.} `+' indicates adding the current method based on the previous line. We report the results of linear evaluation on ImageNet.}
\label{tab:ab1}
\scalebox{1.0}{
\begin{tabular}{l|l|l|l|l}
\hline
\diagbox{Method}{Pre-epochs}    & 100 & 200 & 300 & 800  \\ \hline
baseline & 54.8\% & 58.8\% & 61.5\% & 64.4\%  \\ \hline
+ parallel mask strategy & 60.0\%  & 63.4\%& 65.3\%  & 68.4\%  \\ \hline
++ self-consistency & \textbf{60.9\% } & \textbf{65.0\%}& \textbf{68.4\%}  & \textbf{70.4\% } \\
\hline
\end{tabular}
}
\end{table}

\textbf{Effect of whole data utilization.} In Section \ref{sec:pa_mask}, we have discussed the superiority of the parallel mask strategy we proposed. As shown in Table \ref{tab:whole_data}, we show four designs of data utilization. Table \ref{tab:whole_data} (a) presents the result of 200-epoch MAE with $1$ time purely random masking from train images in each iteration, and it is regarded as the baseline. Then, Table \ref{tab:whole_data} (b) represents the result of 200-epoch MAE with $4$ times purely random masking from train images in each iteration according to the pure random mask strategy of MAE. Table \ref{tab:whole_data} (c) shows the result of 200-epoch MAE with $4$ times disjoint random masking from train images in each iteration by the parallel mask strategy. Compared with Table \ref{tab:whole_data} (a), the result of Table \ref{tab:whole_data} (b) indicates that sampling multiple times can improve data utilization to a certain extent and enhance the performance of the pre-trained model. However, the pure random mask strategy can not guarantee the utilization of whole data ($100\%$ data utilization) and the sampling times of each patch are different in $4$ masking times. It is worth noting that the parallel mask strategy of Table \ref{tab:whole_data} (c) can theoretically address these problems, and the experimental result of (c) also proves this point: the parallel mask strategy outperforms the pure random mask by $0.9\%$ under the same epochs. Therefore, our strategy can further enhance data utilization and achieve $100\%$ of data utilization rate, thereby improving pre-training efficiency. The effectiveness of our strategy has been validated both theoretically and experimentally. Here, we design an additional experiment called the complementary mask strategy, which involves sampling train images twice. The first randomly sampling  $25\%$ of the data from train images, while the second sampling utilizes the remaining  $75\%$ of train images. This approach also allows the model to utilize the whole data, and the computational complexity is even higher than that of the parallel mask strategy due to the computational characteristics of transformers. However, the complementary mask strategy ultimately exhibits performance that is even worse than the baseline. This is because the $75\%$ part represents the $25\%$ mask ratio of the pre-training process, the low mask ratio leads to the worse transfer ability according to the prior work \cite{he2021masked}. Therefore, we adopt the parallel mask strategy as our design of whole data utilization.

\begin{figure}[t]
    \centering
    \includegraphics[width=1.0\linewidth]{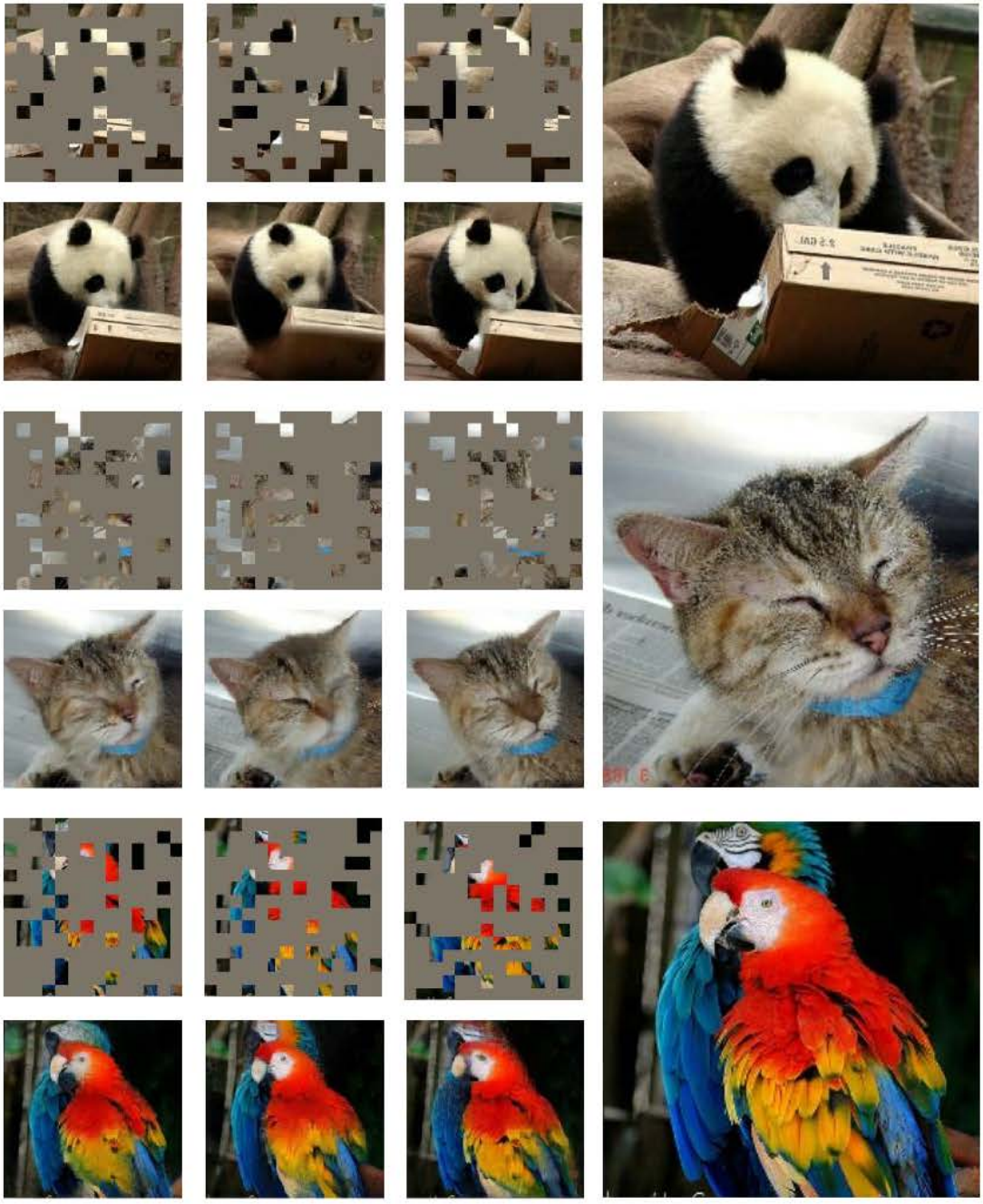}
    \caption{
        \textbf{Different reconstruction results of our EMAE correspond to different mask seeds.} Different combinations of visible patches are sampled from the same image by the random seeds, then these combinations are fed into our EMAE and EMAE generates the reconstructed images. Here, we reconstruct three images as examples. These reconstruction results contain similar semantics and become matched closely with each other, demonstrating the effectiveness of our self-consistency learning.
    }
    \label{fig:figure2}
\end{figure}

\textbf{Effect of self-consistency learning.} To further improve consistency, we propose the self-consistency learning to encourage the model to generate reliable representations in the pre-training process. On the basis of the whole data utilization design, the self-consistency loss is introduced in the pre-training stage. As shown in Table \ref{tab:self-con}, Table \ref{tab:self-con} (a) presents the result of the parallel mask strategy on MAE, which is the pixel reconstruction task. Furthermore, Table \ref{tab:self-con} (b) exhibits the result of pure self-consistency learning without pixel reconstruction based on the parallel mask strategy. The result indicates that the pure self-consistency model without pixel-level reconstruction can still learn impressive semantic features, and its performance even outperforms that of pixel reconstruction.  Table \ref{tab:self-con} (c) shows the result of self-consistency learning with pixel reconstruction, and its performance slightly outperforms pure self-consistency learning. Since pure self-consistency learning loses the generation and reconstruction capabilities, it cannot be extended to the image synthesis task \cite{li2023mage,wei2023diffusion}. Therefore, we still retain the loss for pixel reconstruction and adopt design (c) as the final solution. Notably, after self-consistency learning is introduced to the model, it can be observed that the reconstructed images generated by our EMAE end up matching closely each other in Figure \ref{fig:figure2}. From the above description, it illustrates the effectiveness and superiority of the self-consistency mechanism, which further improves the performance and efficiency of MIM.

\begin{table}[!t]
\centering
\caption{\textbf{Ablations for EMAE: Effect of longer pre-training epochs.} We report the results of linear evaluation on ImageNet.}
\label{tab:longer}
\scalebox{1.1}{
\begin{tabular}{l|c|c|c}
\hline
Metrics & Pre-train data  &  Pre-train epochs & Top-1  \\ \hline
(a) linear probing & \multirow{2}{*}{ImageNet-$1$K} & \multirow{2}{*}{1600} & 71.2\%  \\ \cline{0-0} \cline{4-4}
(b) fine-tuning  & &   & 84.2\%  \\ 
\hline
\end{tabular}}
\end{table}


\textbf{Effect of the proposed method on different pre-training epochs.} As shown in Table \ref{tab:ab1}, the second line shows the results of MAE pre-training on the ImageNet dataset for different pre-training epochs ($e.g.$, $100$, $200$, $300$, and $800$ pre-training epochs) as our baseline.
The third line indicates that the whole data is divided into $4$ non-overlap parts under different pre-training epochs.  Compared with the baseline, the results of the third line surpass the MAE by around $4.0 \% \sim 5.2 \%$ with the same pre-training epochs, demonstrating that sufficient training data can effectively improve performance on different pre-training epochs. Moreover, these results illustrate the importance of exploiting the whole data, which can efficiently increase data utilization and improve pre-training efficiency. Based on the design of the whole data utilization, the results of self-consistency learning are listed in the fourth line of Table \ref{tab:ab1}. Compared with the whole data utilization design, the results of the fourth line further enhance the performance of the pre-trained model, $e.g.$, the results of the fourth line surpass that of the third line by around $0.9 \% \sim 2.9 \%$ with the same pre-training epochs. Hence, EMAE outperforms the MAE by around $6.0\% \sim 6.7\%$ with the same pre-training epochs. 

\textbf{Effect of longer pre-training epochs.} To further investigate the performance of EMAE for longer pre-training epochs, we train EMAE with ViT-B on ImageNet-$1$K for $1600$ epochs. Then, the pre-trained model is loaded in linear probing and end-to-end fine-tuning classification tasks. Ultimately, its result in linear probing is $71.2\%$ and the fine-tuning result is $84.2\%$. It shows our EMAE can still obtain gains from longer pre-training epochs.

\textbf{Effect of the $K$ division.} According to Section \ref{sec:pa_mask}, the $K$ directly determines the mask ratio of each part, and the mask ratio is $\frac{K-1}{K}$. In Table \ref{tab:ab2}, it can be observed that our approach constantly surpasses the MAE with different high mask ratios ($e.g.$, $\frac{3}{4}$, $\frac{6}{7}$, and $\frac{13}{14}$ mask ratios) in the same pre-training settings. Meanwhile, the performance of our approach acquires the best when the $K$ is set to be $4$ (the mask ratio is $75\%$), and the phenomenon also fits the observation of MAE about the mask ratio.

\begin{table}[!t]
\centering
\caption{\textbf{Ablations for EMAE: Effect of the $K$ division.} We report the results of linear evaluation on ImageNet.}
\label{tab:ab2}
\scalebox{1.2}{
\begin{tabular}{l|c|c|c}
\hline
\diagbox{Method}{Pre-epochs}    & 100  & 300 & 800  \\ \hline
(a) MAE mask ratio = $\frac{3}{4}$ & 54.8\%  & 61.5\% & 64.4\%  \\ \hline
(b) EMAE K = 4 & 60.9\%  & 68.4\% & 70.4\%  \\ \hline
(c) MAE mask ratio = $\frac{6}{7}$ & 53.3\% & 61.0\%  & 63.9\%\\ \hline
(d) EMAE K = 7 & 60.5\% & 66.5\%  &  68.1\% \\  \hline
(e) MAE mask ratio = $\frac{13}{14}$  & 46.4\% &  54.7\% & 60.7\% \\ \hline
(f) EMAE K = 14 & 52.5\% & 61.0\%  &  66.7\% \\ 
\hline
\end{tabular}
}
\end{table}

\begin{table}[!t]
\centering
\caption{\textbf{Ablations for EMAE: Reliability of the generations.} We report the results of linear evaluation on ImageNet.}
\label{tab:generations}
\scalebox{1.3}{
\begin{tabular}{l|c|c}
\hline
Metrics & Validation data   & Top-1  \\ \hline
\gray{(a) DINO v2} &\multirow{3}{*}{ImageNet-$1$K \texttt{val} set}& \gray{86.6\%} \\ \cline{0-0} \cline{3-3}
(b) MAE &   & 80.4\%  \\ \cline{0-0} \cline{3-3}
(c) EMAE  &   & 86.2\%  \\ 
\hline
\end{tabular}}
\end{table}

\begin{table}[t]
    \centering
    \caption{ \textbf{Ablations for EMAE: Image generation criteria.} We report SSIM~\cite{wang2004image}, LPIPS~\cite{zhang2018unreasonable}, and FID~\cite{heusel2017gans} results. The architecture of various methods adopts ViT-L. 
    } 
    \label{tab:quality}
    \setlength{\tabcolsep}{5pt}
\renewcommand{\arraystretch}{1.15}
    \footnotesize
    \scalebox{1.15}{
        \begin{tabular}{l c  c c}
        \toprule
            Method   & SSIM$\uparrow$ & LPIPS$\downarrow$ & FID$\downarrow$
              \\
        \midrule
        MAE & 0.668& 0.400&48.8 \\
        EMAE & 0.734 &0.296 & 29.7 \\
        \bottomrule     
    \end{tabular} }
\end{table}

\textbf{Image generation quality with EMAE.} As shown in Figure~\ref{fig:figure2}, the reconstructions by EMAE are semantically consistent with the original image, demonstrating the reliability of reconstructions. To further substantiate the reliability of EMAE, we conduct experiments by comparing the classification performance by feeding images generated separately by MAE and EMAE into a classifier. Specifically, we use a random mask with a mask ratio of $75\%$ on ImageNet \texttt{val} set to reconstruct images by MAE and EMAE, respectively. These images are then fed into a SOTA image classifier, DINO v2 with ViT-g \cite{oquab2023dinov2}, to perform the linear probing task. The classification results are presented in Table \ref{tab:generations}. Compared with the classification result of the original images on DINO v2, the reconstructions by MAE show an obvious performance drop in the classification task, whereas the ones by EMAE are remarkably better than MAE in the classification task and achieve comparable performance to the original images. It demonstrates that EMAE alleviates the unreliability and inconsistency problem. Moreover, we further validate the image generation quality of EMAE, and the final results are presented in Table \ref{tab:quality}. Here, SSIM~\cite{wang2004image} and LPIPS~\cite{zhang2018unreasonable} are standard criteria to measure the similarity between generated images and the original, while FID~\cite{heusel2017gans} is standard metric to measure the fidelity and diversity of the generated images. EMAE demonstrates superior performance to MAE across all three criteria. However, there remains a substantial performance drop compared to dedicated image generation models. Meanwhile, extensions of MAE, such as MAGE \cite{li2023mage} and RCG \cite{li2023self}, have already achieved SOTA performance in image generation. It's worth noting that our method is orthogonal to these approaches. In the future, we plan to explore the effectiveness of our method in the context of image generation.

\section{Conclusion}

In this paper, we investigate the two serious problems caused by the high mask ratio of MIM, namely prolonged pre-training and the high inconsistency of the pre-trained model. To overcome the above problems, we propose an approach, called efficient masked autoencoders with self-consistency (EMAE). To improve the data utilization in the self-supervised task, EMAE proposes the parallel mask strategy to achieve $100\%$ of data utilization rate, thus enhancing the pre-training efficiency. At the same time, self-consistency learning is proposed to decrease the uncertainty and inconsistency of MIM, which further improves performance. After pre-training on various datasets, the proposed EMAE shows good versatility and scalability in multiple downstream visual tasks, such as linear evaluation, finetuning classification, object detection, instance segmentation, and semantic segmentation. We expect that our study can attract the community’s attention to more efficient and reliable MIM.
Additionally, this work also has some limitations: 1) We validate the performance of EMAE by constructing experiments on the ImageNet, COCO, and OpenImages datasets. However, the promise of self-supervised learning is to establish a general feature extractor with larger datasets. We have not extended this method to larger datasets \cite{zhou2017places,thomee2016yfcc100m,sun2017revisiting} and larger architectures ($e.g.$, ViT-H) due to the resource and time consumption. 2) Meanwhile, the self-consistency of EMAE has the semantic understanding for input visual patches but its reconstructions of individual images under high mask conditions may still not be that perfect. This problem may be solved by the choice of self-consistency loss or by employing generative methods for assistance \cite{li2023mage,wei2023diffusion}. These limitations warrant further research and consideration when building upon this work to achieve a better self-supervised learning method.

\section*{Acknowledgments}

This work was supported by National Key R\&D Program of China under Grant No.2021ZD0110400 and National Natural Science Foundation of China (No.62276260, No.62176254, No.62076235).

\bibliographystyle{IEEEtran}
\bibliography{egbib}

\newpage

\section{Biography Section}

\begin{IEEEbiography}[{\includegraphics[width=1in,height=1.25in,clip,keepaspectratio]{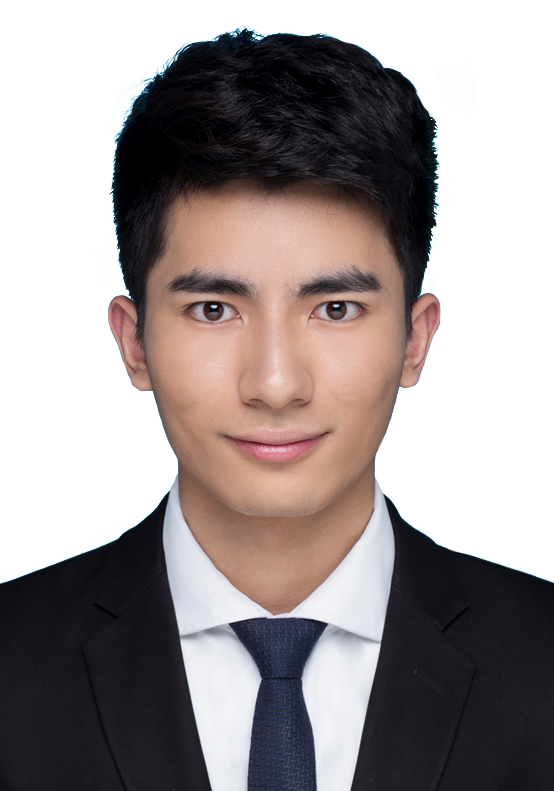}}]{Zhaowen Li}
received the B.E. degree from University of Science and Technology Beijing, Beijing, China, in 2019. He is currently pursuing the Ph.D. degree in pattern recognition and intelligence systems with the Foundation Model Research Center, Institute of Automation, Chinese Academy of Sciences. His current research interests include pattern recognition and machine learning, self-/un-supervised learning and computer vision, and multimodal learning.
\end{IEEEbiography}

\begin{IEEEbiography}[{\includegraphics[width=1in,height=1.25in,clip,keepaspectratio]{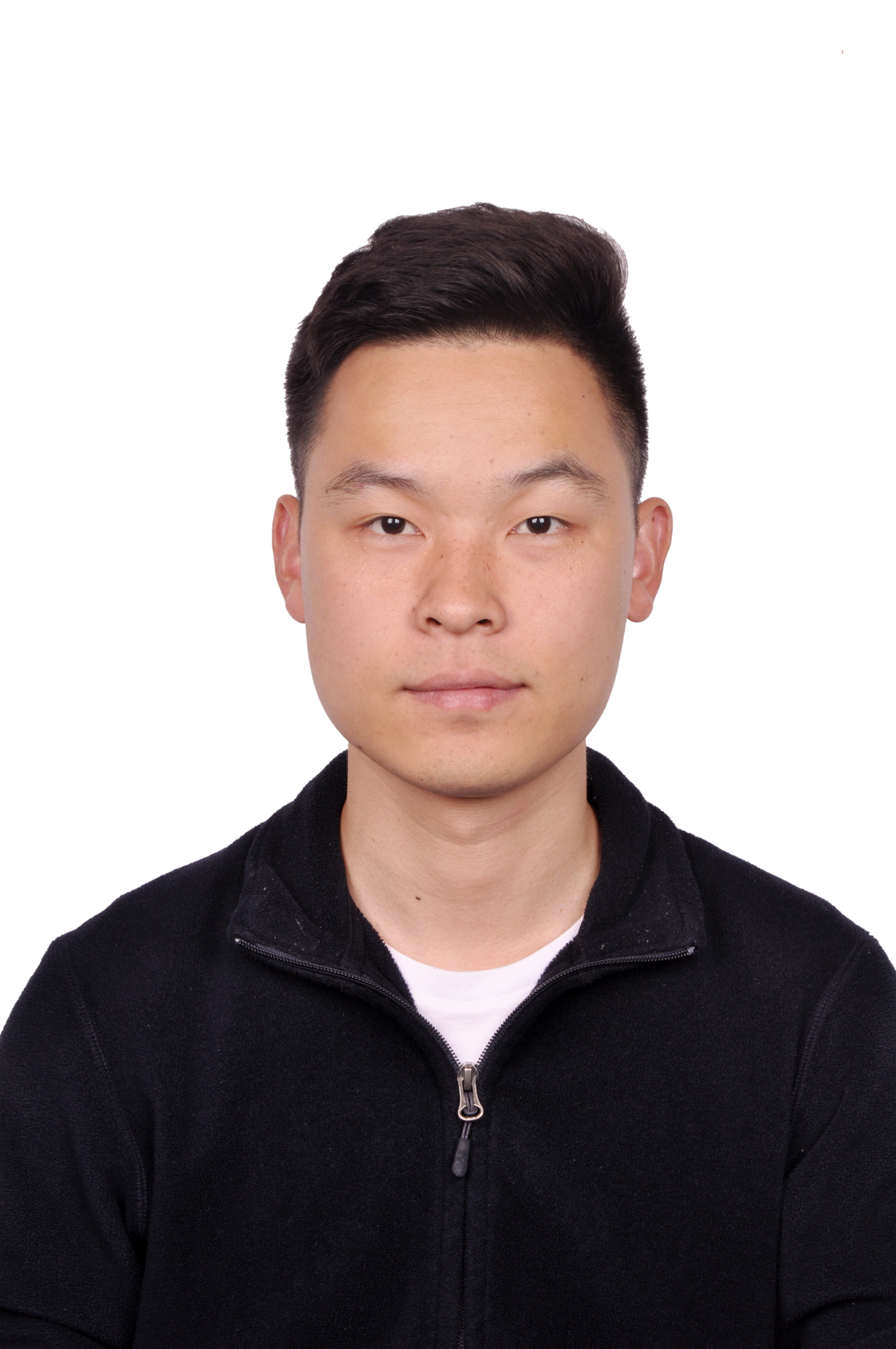}}]{Yousong Zhu} 
received the B.E. degree in 2014 from Central South University, Changsha, China and the Ph.D. degree in pattern recognition and intelligence systems from the Institute of Automation, Chinese Academy of Sciences, in 2019. He is currently an Associate Researcher with the Foundation Model Research Center, Institute of Automation, Chinese Academy of sciences. His current research interests include pattern recognition and machine learning, computer vision and object detection, and self-supervised learning and general vision model.
\end{IEEEbiography}

\begin{IEEEbiography}[{\includegraphics[width=1in,height=1.25in,clip,keepaspectratio]{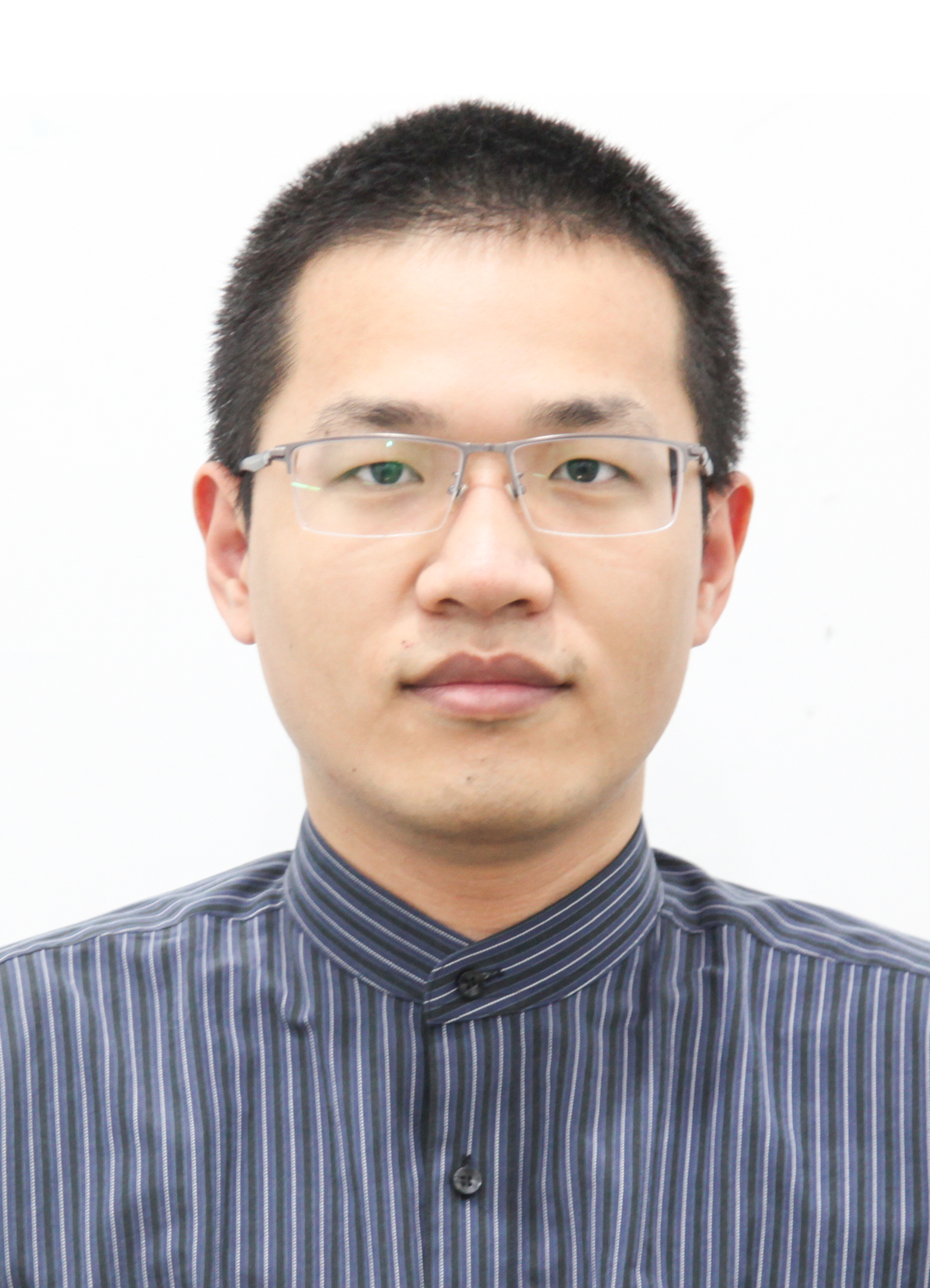}}]{Zhiyang Chen}
received the B.E. degree from Xi’an Jiaotong University in 2019. He is currently pursuing the Ph.D. degree in pattern recognition and intelligence systems with the Foundation Model Research Center, Institute of Automation, Chinese Academy of Sciences. His current research interests include pattern recognition and machine learning, computer vision and object detection.
\end{IEEEbiography}

\begin{IEEEbiography}[{\includegraphics[width=1in,height=1.25in,clip,keepaspectratio]{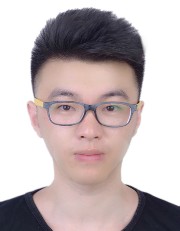}}]{Wei Li}
received the M.Sc. Degree from National Chiao Tung University, Hsinchu, Taiwan, in 2019.
He is currently a Senior Researcher in  SenseTime Smart City Group. His research focuses on large models in computer vision and their applications, especially in semi-supervised, self-supervised representation learning, multimodal large language models, and industrial defect detection.
\end{IEEEbiography}

\begin{IEEEbiography}[{\includegraphics[width=1in,height=1.25in,clip,keepaspectratio]{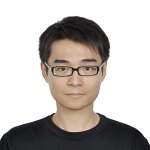}}]{Rui Zhao}
received his Bachelor’s degree from University of Science and Technology of China (USTC) in 2010, his Ph.D. degree from The Chinese University of Hong Kong (CUHK) in 2015. He is a leading expert in AI for face recognition and person re-identification. He has gained over 5,000 Google Scholar Citations with recent 20 publications during last few years. He is currently an Executive Research Director at SenseTime and head of Shared Technology Center (STC) of R$\&$D; in SenseTime Smart City Group (SCG), where he leads a team of talented and energetic researchers and engineers to foster and develop innovative techniques for smart city solutions. His research interests include deep learning, machine learning, and computer vision.
\end{IEEEbiography}

\begin{IEEEbiography}[{\includegraphics[width=1in,height=1.25in,clip,keepaspectratio]{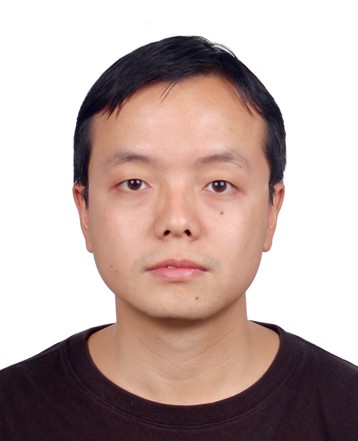}}]{Chaoyang Zhao}
received the B.E. degree and the M.S. degree from University of Electronic Science and Technology of China, Chengdu, China, in 2009 and 2012, respectively, and the Ph.D. degree in pattern recognition and intelligent system from the Chinese Academy of Sciences, Beijing, China, in 2016. His current research interests include pattern recognition and machine learning, image and video processing and intelligent video surveillance.
\end{IEEEbiography}

\begin{IEEEbiography}[{\includegraphics[width=1in,height=1.25in,clip,keepaspectratio]{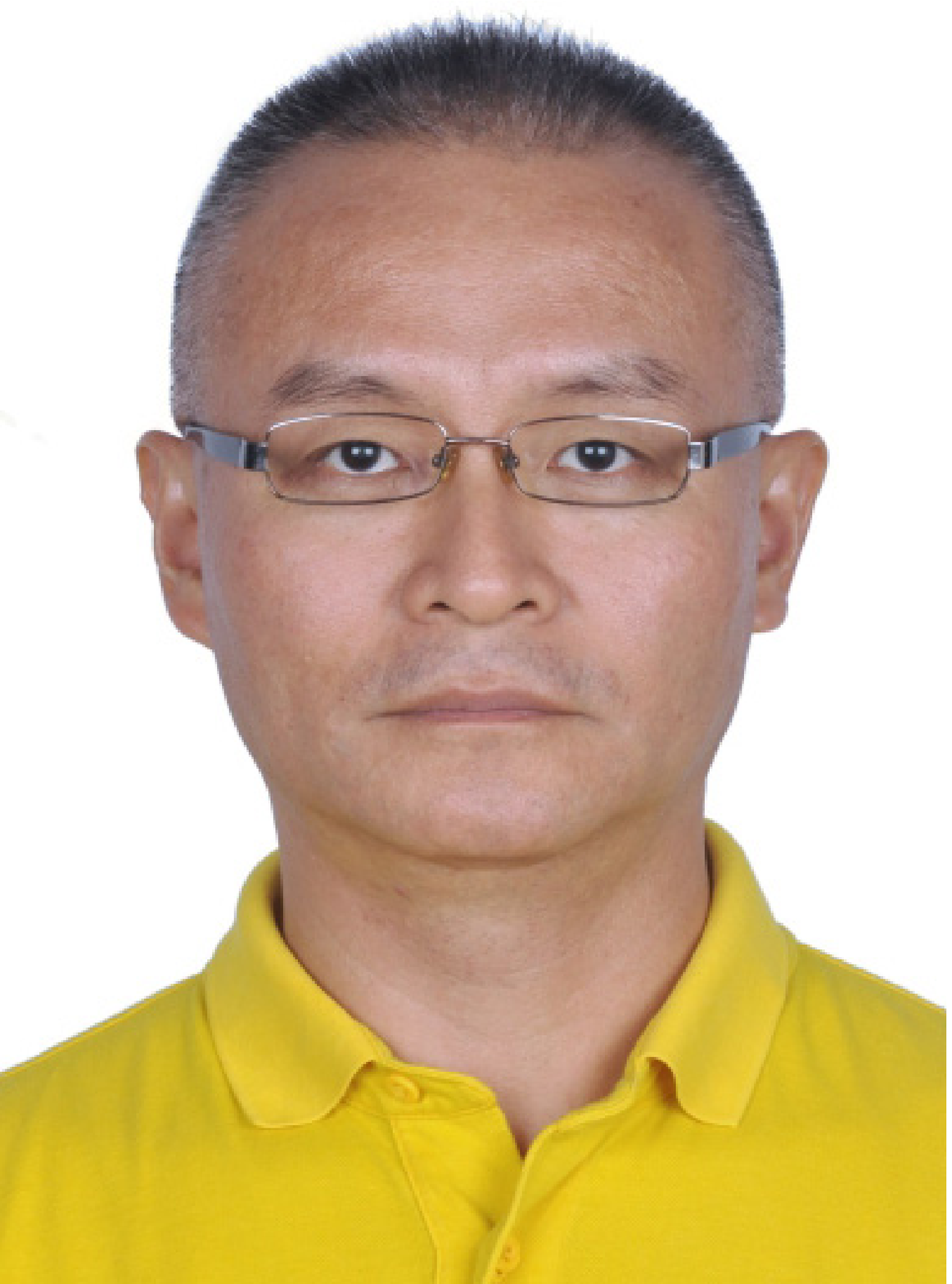}}]{Ming Tang}
received the B.S. degree in computer science and engineering and M.S. degree in artificial intelligence from Zhejiang University, Hangzhou, China, in 1984 and 1987, respectively, and the Ph.D. degree in pattern recognition and intelligent system from the Chinese Academy of Sciences, Beijing, China, in 2002. He is currently a Professor with the Foundation Model Research Center, Institute of Automation, Chinese Academy of Sciences. His current research interests include computer vision and machine learning.
\end{IEEEbiography}

\begin{IEEEbiography}[{\includegraphics[width=1in,height=1.25in,clip,keepaspectratio]{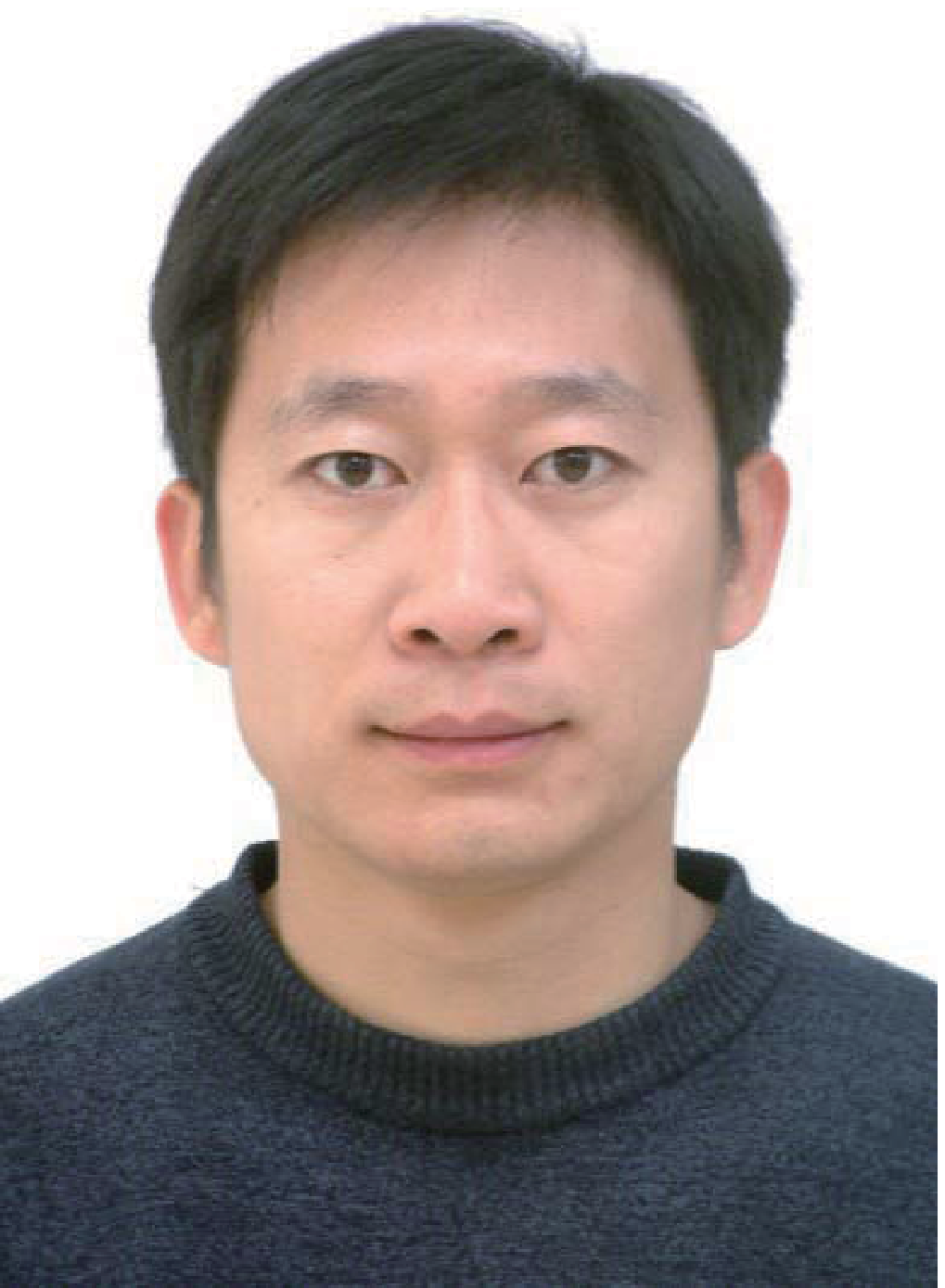}}]{Jinqiao Wang}
received the B.E. degree from the Hebei University of Technology, China, in 2001, the M.S. degree from Tianjin University, China, in 2004, and the Ph.D. degree in pattern recognition and intelligence systems from the Foundation Model Research Center, Chinese Academy of Sciences, in 2008. He is currently a Professor with the Chinese Academy of Sciences. His research interests include pattern recognition and machine learning, image and video processing, mobile multimedia, and intelligent video surveillance.
\end{IEEEbiography}

\vfill

\end{document}